\documentclass[11pt]{article}

\usepackage[preprint]{acl}

\usepackage{times}
\usepackage{latexsym}
\usepackage{multirow}
\usepackage{amsmath}
\usepackage{booktabs}
\usepackage{amssymb}

\usepackage[T1]{fontenc}
\usepackage[utf8]{inputenc}

\usepackage{microtype}

\usepackage{inconsolata}

\usepackage{graphicx}

\newcommand{\sysname}{\textsc{P2C}}

\title{Read Before You Fit: Text-Derived Constraints for Discovering\\ Differential Equations from Noisy Biological Data}

\author{Pronoma Banerjee \\
  Purdue University \\
  \texttt{baner102@purdue.edu} \\\And
  Md. Masudur Rahman \\
  Purdue University \\
  \texttt{rahman64@purdue.edu} \\\And
  Juan P. Wachs \\
  Purdue University \\
  \texttt{jpwachs@purdue.edu} \\}

\begin{document}
\maketitle
\begin{abstract}
Large language models are increasingly used for scientific equation discovery. However, most existing methods are strongly data-driven and assume sufficiently clean and representative observations. This assumption is problematic in biological systems, where individual-level measurements are often sparse, noisy, heterogeneous, and expensive to obtain. While the underlying mechanisms are extensively described in scientific text, they are not explicitly used for the LLM based modeling of such systems. We introduce P2C, an agentic framework that translates textual descriptions of biological systems into a typed set of formal constraints, and use these constraints to restrict candidate ordinary differential equation (ODE) structures before numerical fitting. P2C consists of five stages: Text2Constraint extraction, ODE discovery, symbolic verification, parameter fitting, and model scoring. We evaluate P2C on four published ODE systems of acute inflammation and immune regulation, under clean and noisy observations. By progressively degrading the numerical evidence, we isolate when textual domain knowledge contributes beyond regression alone. Across the four systems, adding Text2Constraint reduces the worst form error from 1.397 to 0.499 on clean data and from 13.096 to 1.095 for data perturbed with Gaussian noise of \(\sigma=0.1\); P2C also achieves lower form error and trajectory error than several LLM-based baselines over the tested noise range. \end{abstract}

\section{Introduction}

Differential equations are widely used to describe the dynamics of biological systems, including immune and infectious-disease processes \cite{Vandenbon2012TLR}, inflammatory responses \cite{Kumar2004Dynamics}, and biochemical reactions. However, complex biological mechanisms are often communicated through natural-language descriptions, pathway diagrams, and qualitative ``word models'' rather than complete mathematical formulations \cite{gyori2017}.

Symbolic regression (SR) seeks to recover interpretable equations directly from data \citep{schmidt2009distilling,brunton2016sindy}, making it attractive for mechanistic biological modeling \citep{mangan2016sindybiological}. Yet most equation-discovery methods remain strongly data-driven and assume that observations are sufficiently clean and informative to identify the governing dynamics. This assumption is often unrealistic in biology, where measurements may be sparse, noisy, heterogeneous, and expensive to obtain. As a result, multiple equations can fit the same observations while encoding very different mechanisms, particularly when models are extrapolated beyond the observed regime \cite{Gutenkunst2007Sloppy, Raue2009Identifiability}.

Biology nevertheless provides an additional source of information: mechanistic knowledge expressed in scientific text. The literature frequently specifies whether one process activates or inhibits another, whether responses saturate, whether certain interactions are absent, and which quantities must remain positive. Recent work has incorporated domain knowledge and physics-informed structure into symbolic regression \cite{machado2023sparseidentificationnonlineardynamics, Taskin2026Knowledge, SimonKeren2023}, while LLM-SR \cite{shojaee2025llmsr} showed that large language models can provide scientific priors and programmatic guidance during symbolic search. However, when textual descriptions are provided directly to an LLM, it is difficult to determine what information was actually extracted from the text or whether successful equation recovery was driven primarily by numerical fitting.

% We introduce \sysname{} (Prose-to-Constraints), a multi-agent framework that makes the contribution of scientific language explicit. Instead of directly prompting an LLM to generate an equation from prose and data, \sysname{} first translates the prose description into a typed set of formal constraints, including interaction sign, saturation, positivity, and structural-absence constraints. Candidate ODE structures are then generated and symbolically verified against these constraints before numerical parameter fitting and model scoring. This intermediate representation makes the linguistic contribution inspectable and allows failures in knowledge extraction to be separated from failures in downstream equation search.

We introduce \textbf{P2C (Prose-to-Constraints)}, an agentic framework
that makes the contribution of scientific language explicit and
verifiable. Rather than using biological prose only as a prompt, P2C
first translates it into a typed set of formal predicates describing
properties such as interaction signs, saturation, structural absence,
joint interactions, source terms, and positivity. Each extracted
constraint is grounded to a source span and checked for entailment and
consistency. Candidate ODE structures are then tested against these
predicates before numerical parameter fitting. This separates
what is inferred from language from what is inferred
from numerical observations, and allows failures in knowledge
extraction, symbolic structure discovery, and parameter estimation to
be analyzed separately.

We evaluate P2C on four published models of acute inflammation and immune regulation \cite{Reynolds2006Reduced, Kumar2004Dynamics, Dunster2014Resolution, Baker2013CytokineInflammation} under clean and degraded numerical observations. The results show a consistent descriptive trend as the numerical
evidence weakens: P2C obtains the lowest median form error on three
of four systems at $\sigma=0.05$ and on all four systems at
$\sigma=0.1$. We additionally apply
P2C to measured dose-response data from splenic focused-ultrasound
treatment of inflammatory arthritis \cite{Zachs2019Noninvasive}, testing whether mechanistic knowledge can restrict equation forms when the available observations
alone do not determine biologically plausible extrapolation.

Our main contributions are:

\begin{enumerate}
    \item We introduce \textbf{P2C}, a modular agentic framework that
    translates biological prose into attributable, typed mathematical
    constraints and uses them to restrict ODE discovery before
    numerical fitting.

    \item We develop an explicit constraint representation and
    symbolic verification procedure for mechanistic properties
    including interaction sign, saturation, structural absence,
    joint interaction, source, and positivity constraints, making the
    linguistic contribution to equation discovery inspectable and
    independently verifiable.

    \item We construct a controlled evaluation suite of four published
mathematical-immunology models paired with literature-derived
descriptions and sentence-level reference constraints, and evaluate
P2C under progressively degraded observations.

    \item We demonstrate the framework on measured inflammatory-
    arthritis data from splenic focused ultrasound, where P2C
    recovers a bounded intervention-response equation with lower
    held-out prediction error and fewer mechanistic constraint
    violations than the evaluated baselines.
\end{enumerate}

\section{Problem Formulation}
We consider a multimodal equation discovery task where the input consists of 3 parts: (i) the numerical time-series observation $\mathcal{D} = \{(t_i, x(t_i))\}$ (clean data or data with Gaussian Noise $\mathcal{N}(0, \sigma)$), (ii) a prose description of the system $\mathcal{T}$ (a mechanism or pathway) and (iii) state variables for the system and their corresponding words in the prose. Our goal is to discover interpretable and biologically plausible system of ODEs: $\dot{x}(t) = f(t, x)$, where $f = [f_0, \cdots, f_{d-1}]^T$ is composed of functional terms proposed by an LLM. 

The prose description $\mathcal{T}$ can be mapped to a set of formal
predicates $\mathcal{C}(\mathcal{T})=\{c_1,\ldots,c_K\}$, each constraining the admissible structure of the vector field ${f}$. These predicates define the feasible hypothesis class
\begin{equation}
    \mathcal{F}_{\mathcal{T}}
    =
    \left\{
    f\in\mathcal{F}:
    c_k(f)=\mathrm{True},
    \ \forall c_k\in\mathcal{C}(\mathcal{T})
    \right\}.
\end{equation}
Equation discovery therefore seeks
\begin{equation}
    (\hat{f},\hat{\boldsymbol{\theta}})
    =
    \arg\min_{f\in\mathcal{F}_{\mathcal{T}},
              \boldsymbol{\theta}\in\Theta}
    \mathcal{L}_{\mathrm{data}}
    (f,\boldsymbol{\theta};\mathcal{D}),
\end{equation}
so that textual knowledge restricts the space of admissible ODE
structures while the numerical observations determine the fitted
model within this constrained space.

\section{Method}
\label{sec:methods}

\begin{figure*}[t]
    \centering
    \includegraphics[width=1\linewidth]{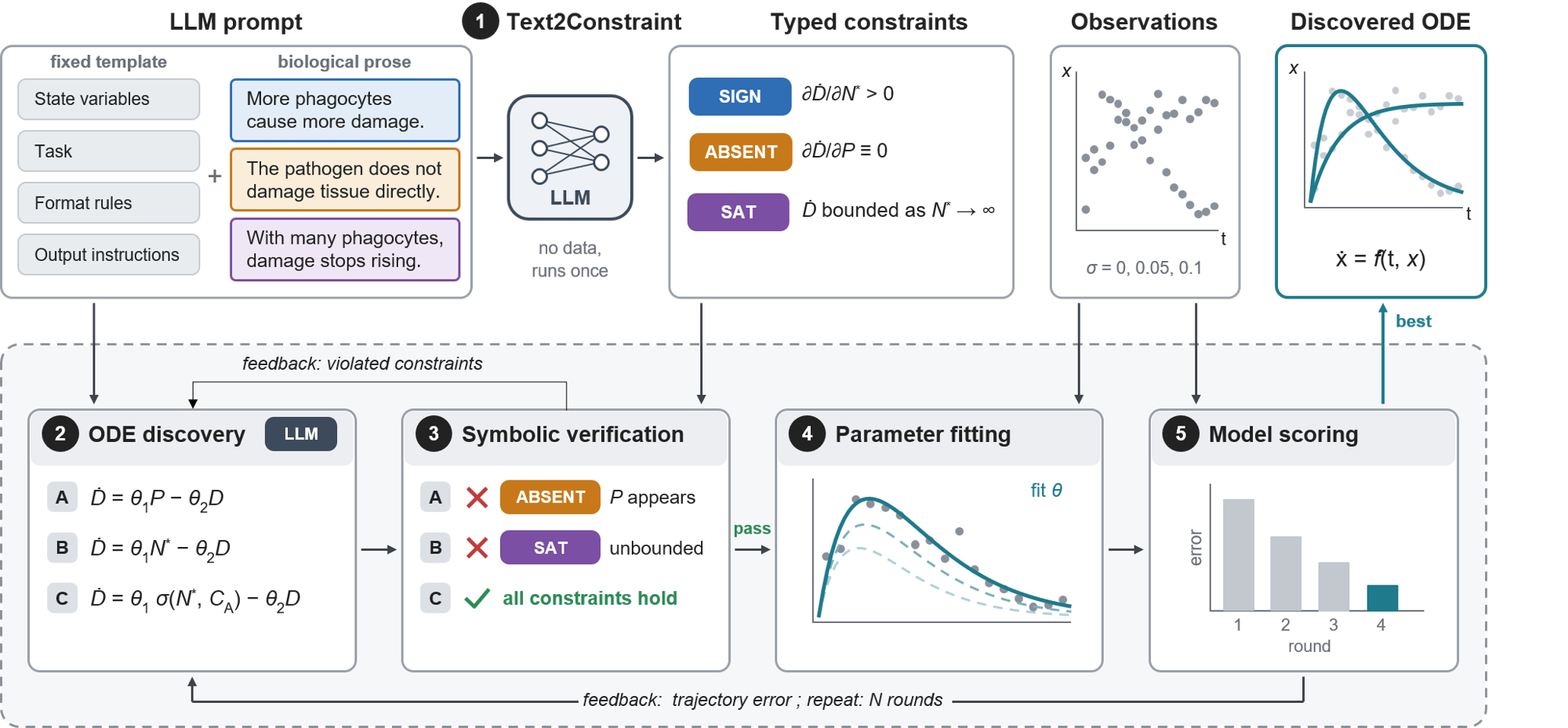}
    \caption{Overview of the P2C framework for the LLM-based ODE discovery under text-derived formal constraints. \textbf{(1) Text2Constraint stage:} An LLM converts each sentence of the biological description into verifiable constraints. \textbf{(2) ODE discovery stage:} In each round, an LLM proposes candidate systems from the variable list, the description, and feedback from earlier rounds, \textbf{(3) Symbolic verification stage:} The verifier rejects every candidate that violates a constraint before any fitting. If all proposed equations violate constraints, go back to Stage 2. \textbf{(4) Parameter fitting stage:} The parameters of the remaining candidates are fitted to observations. \textbf{(5) Model scoring stage:} Each fitted candidate is scored by its trajectory error on held-out validation points of the same noisy observations, and the best candidates and their scores are fed back. Stages 2--5 repeat for $N$ rounds; and the best model over all rounds is saved.}
    \label{fig:pipeline}
\end{figure*}

The P2C pipeline is shown in Figure \ref{fig:pipeline}. It first converts a textual description of a biological system into explicit constraints, then proposes and verifies candidate ODEs, fits their parameters to observations, and selects a model by its held-out trajectory error. The pipeline never uses the ground-truth equations to propose, fit, or select candidates.
 
\begin{table}[t]
\centering\small
\begin{tabular}{@{}ll@{}}
\toprule
\textbf{Type} & \textbf{Predicate on $f_i$} \\
\midrule
\textsc{Sign}   & $\partial f_i/\partial x_j > 0$ (or $<0$) $\forall x \in \mathbb{R}^d_{\ge 0}$ \\
\textsc{Absent} & $\partial f_i/\partial x_j \equiv 0$ \\
\textsc{Sat}    & $\lim_{x_j\to\infty} f_i$ is finite \\
\textsc{Inter} & $\exists\ f_i \in f
\text{ such that } x_j,x_k \in f_i$
\\
\textsc{Source} & $f_i(t, \mathbf{0}) > 0$ \\
\textsc{Pos}    & $x_i = 0 \Rightarrow f_i \ge 0$ \\
\bottomrule
\end{tabular}
\caption{Different types of constraints obtained from the prose descriptions, and checked by Stage 3.}
\label{tab:schema}
\end{table}
 
% \paragraph{(1) Text2Constraint.}
% An LLM receives $\mathcal{T}$, the glossary, and the definitions in
% Table~\ref{tab:schema}, and returns a list of typed constraints. It is
% instructed to use only what $\mathcal{T}$ states, to emit \textsc{Absent}
% only when an interaction is explicitly denied, and to attach to every
% constraint the sentence that supports it. The supporting span makes each
% extracted constraint auditable and lets us score extraction against gold
% annotations independently of equation discovery. This stage sees no data and
% runs once; the resulting set $\mathcal{C}$ is not revised afterwards.

\paragraph{Stage 1: Text2Constraint extraction and grounding.} Instead of using natural language as a loose prompt, as in several prior LLM based equation discovery approaches \cite{song2026discovering, holt2024datadriven}, P2C translates the biological description $\mathcal{T}$ into a strict set of typed constraints $\hat{C}$. These constraints are described in Table \ref{tab:schema}. To prevent hallucinations or incorrect translations, every extracted constraint passes through 4 rigorous grounding checks before being appended to the final contraint set: (1) Attribution: Every constraint must be tied to a corresponding substring from the source text. (2) Entailment: A separate verifier LLM checks whether the verbatim substring logically entails the verbalized constraint template. (3) Self-Consistency: The constraint must be consistently predicted across a majority of $K$ independent extraction runs. (4) Logical Consistency: The final constraint set $\hat{C}$ is filtered to remove contradictory predicates (e.g., mutually exclusive signs for the same state variables). 
 
\paragraph{Stage 2: LLM-Driven ODE Discovery}
An LLM generator proposes $H$ candidate equation structures per round. To constrain the functional space, candidates are formulated as sums of at most $K$ terms featuring positive symbolic parameters $\theta$, explicitly defined signs, and integer exponents bounded between 1 and 6. To enforce a strict separation between hypothesis generation and verification, the generator's prompt includes the system description, the state variable, and iterative feedback with violated constraints, but is deliberately blinded to both the numerical observations and the formally extracted constraint set $\hat{C}$.
 
\paragraph{Stage 3: Symbolic verification.}
Each candidate ODE is parsed as a symbolic expression and tested against every constraint in $\hat{C}$ using the \texttt{SymPy} symbolic mathematics library \cite{meurerSympy}. We use symbolic differentiation and simplification for SIGN and ABSENT constraints, symbolic limits for SAT, and symbolic substitution for SOURCE and POS. All state variables are assumed non-negative and all free parameters positive. When the sign of a partial derivative cannot be established symbolically, we evaluate it at 400 random points with positive states and parameters and accept it only if the required sign is uniform; this provides numerical evidence rather than a proof. A candidate that violates any constraint is rejected before fitting, and the violated predicate is returned to the proposer. 
 
\paragraph{Stage 4: Parameter fitting.}
For candidates that pass symbolic verification, unknown parameters are fitted independently for each equation. Because direct numerical differentiation severely amplifies measurement noise, P2C employs a weak-form objective, following \cite{Messenger2021WeakSINDy} to
avoid directly differentiating noisy observations (details provided in Appendix \ref{app:weak_form}). Multiplying the ODE by a compactly supported test function $\phi_k$ and integrating by
parts yields
\begin{equation}
    -\int \dot{\phi}_k(t)y_i(t)\,dt
    =
    \int \phi_k(t)
    f_i(\mathbf{y}(t);\boldsymbol{\theta}_i)\,dt ,
\end{equation}
where the integrals are numerically approximated using the datapoints
in $\mathcal{D}_{\mathrm{}}$. The parameters are chosen to minimize the summed squared weak-form residuals across test functions (Appendix~\ref{app:weak_form}). Parameters that enter linearly are  estimated using non-negative least squares, while nonlinear parameters are fitted using bounded optimization with multiple random restarts. 

% Each ODE is fitted separately: other observed state variables may appear on its right-hand side, but the parameters of their governing equations are not optimized jointly. The held-out subset $\mathcal{D}_{\mathrm{eval}}$ is reserved for the subsequent model scoring stage.

\paragraph{Stage 5: Model scoring and Feedback loop.}
Fitted ODEs are scored based on their normalized trajectory error against a held-out validation subset of the noisy observations. Let $\hat x^{r}(t;\hat\theta)$ be the
trajectory of the fitted candidate from the initial state of regime $r$ (representing a specific initial state and set of input parameters), and
$y^{r}_{i}(t)$ the noisy observation of state $i$. The score is the normalised
trajectory error (NMSE) on the validation set,
\begin{equation}
\label{eq:stage5_score}
S=\frac{1}{Rd}\sum_{r=1}^{R}\sum_{i=1}^{d}
\frac{\sum_{t\in V_r}\bigl(\hat x^{r}_{i}(t;\hat\theta)-y^{r}_{i}(t)\bigr)^2}
     {\sum_{t\in V_r}\bigl(y^{r}_{i}(t)-\bar y^{r}_{i}\bigr)^2},
\end{equation}
where $d$ is the number of state variables and $\bar y^{r}_{i}$ is the mean of
$y^{r}_{i}$ over $V_r$. If the candidate cannot be integrated over a regime,
$S=\infty$. 

\paragraph{Feedback and search loop.}
Each scored candidate is added, with its score, to a list $\mathcal{P}$ that is
kept for the whole run. Stage~5 does not accept or reject candidates. In every
iteration Stage~2 receives the $k$ best candidates in $\mathcal{P}$ with their
scores, together with the constraints violated by candidates that Stage~3
rejected in the previous iteration, and proposes new candidates. Each new
candidate is verified, and if it passes it is fitted, scored and added to
$\mathcal{P}$; candidates already in $\mathcal{P}$ are not fitted again. The
loop runs for a fixed number of iterations $N$. 

\paragraph{Final selection.}
After the last iteration, let $S_{\min}$ be the lowest score in $\mathcal{P}$.
Among the candidates with $S\le(1+\tau)\,S_{\min}$, $\tau=0.1$, we return the one with the fewest parameters; ties in
parameter count are broken by the lower validation score $S$.. Because fitting can drive a coefficient to
zero and delete a term that a constraint requires, the selected model is
verified once more with its zero-coefficient terms removed; if it violates a
constraint, the next candidate in the ordering is taken. Only the final system is compared with the ground truth. The values of $k$, $T$ and $\tau$ are
fixed on one development system and are not changed afterwards.

\section{Experiments}\label{sec:experiments}
 
% We organise the experiments around five questions.
% \textbf{RQ1}: does \sysname{} recover the correct form more often than existing
% methods, and does the difference grow as the observations become noisier?
% \textbf{RQ2}: how accurate is Text2Constraint extraction?
% \textbf{RQ3}: is the gain due to the explicit constraints and their symbolic
% verification, or would the same text in a prompt be enough?
% \textbf{RQ4}: what happens when the description is uncertain or wrong?
% \textbf{RQ5}: can recovery be explained by memorisation of published equations?
 
% ---------------------------------------------------------------------
\subsection{Setup}\label{sec:setup}
 
\paragraph{Benchmark and Dataset}
To evaluate our method, we select 4 ODE systems of the inflammatory and immune system, from previously published papers, and name them according to their first author: Dunster (4 variable neutrophil-macrophage interaction model during inflammation regulation) \cite{Dunster2014Resolution}, Baker \cite{Baker2013CytokineInflammation} (2 variable cytokine regulation model in rheumatoid arthritis), Kumar (3 variable pathogen-mediator model) \cite{Kumar2004Dynamics} and Reynolds (4 variable reduced model of acute inflammatory response) \cite{Reynolds2006Reduced}. We supplement each system with a manually created prose description of the pathways/mechanisms, derived from the original papers. Further details are provided in Appendix \ref{sec:app-data}. 
 
\paragraph{Observations and noise.}
For each system we simulate three training regimes and one held-out regime
with different initial states and input values, and sample 201 time points
per regime. State trajectories were generated by numerical integration with LSODA \citep{petzold1983lsoda}, using the Adams method and BDF method, depending on the problem. For the experiments testing the frameworks under noisy data, noise in the form of additive Gaussian, with mean 0 and variance $\sigma\in\{0,0.05,0.1\}$, is applied to each state. Negative values are clipped to zero. All methods receive the same noisy observations. The held-out regime and the noise-free trajectories are used
only for evaluation.
 
% \paragraph{Language models.}
% We run \sysname{} and all LLM-based baselines with \TODO{GPT model name},
% and repeat the main experiment with two open models,
% \TODO{Qwen version} and \TODO{Llama version}. Temperature is \XX.
 
\paragraph{Baseline models}
We evaluate P2C against representative LLM-based equation discovery models via symbolic regression: LLM-SR \cite{shojaee2025llmsr}, DoLQ \cite{song2026discovering} and EDL \cite{10.1063/5.0224297}. These baselines were selected for comparison because of their ability to incorporate textual description of the system $\mathcal{T}$ via prompts and their ability to generate ODE expressions. We use the GPT-6-Luna model \cite{openai2026gpt6} as the default LLM backbone, unless stated otherwise. 
 
% \paragraph{Fair comparison.}
% (i) All LLM-based methods use the same language model, the same description
% and the same budget of \XX{} proposal calls per system.
% (ii) LLM-SR and DoLQ fit equations to time derivatives estimated from the
% data by finite differences, which is sensitive to noise. We therefore run each
% of them in two versions: as published, and with derivatives estimated after
% smoothing \TODO{smoother and its setting}; we report the better of the two and
% give both in Appendix~\ref{app:baselines}. \TODO{check how D3 and MEDA fit
% parameters and state it here.}
% (iii) The Free-text variant shares the fitter and loop of \sysname, so the
% difference between the two isolates the effect of the constraints.
% (iv) Form recovery (Section~\ref{sec:metrics}) re-estimates the parameters of
% every returned model with one common procedure, so it does not depend on the
% fitter of each method.
 
\paragraph{Selected hyperparameters}
\sysname{} runs for $N=25$ iterations with 3 candidates per iteration and
returns the $k=3$ best candidates to the proposal stage. These settings were
fixed on the Reynolds system and not changed afterwards. Each
(system, noise level, method) cell is run with 10 seeds. We report medians form error and trajectory error for each system and each noise level. 
 
% ---------------------------------------------------------------------

\subsection{Evaluation metrics}\label{sec:metrics}
Our primary objective is to evaluate recovery of the \emph{equation
form}, rather than parameter estimation under a particular noise
realization. Let $f^{\star}$ be the ground-truth system, $f$ the
returned model with fitted parameters $\hat\theta$, and $d$ the number of state variables.

\paragraph{Trajectory error.}
We evaluate predictive accuracy using the normalized trajectory error
(NMSE) defined in Eq.~\ref{eq:stage5_score}. During Stage~5, this quantity is computed
against the noisy validation observations $\mathbf{y}^r$ and is denoted
by $S$. For evaluation, we use the same definition but replace the noisy
observations with the corresponding noise-free ground-truth trajectories
$\mathbf{x}^{\star r}$. We report this NMSE on a held-out regime that is
never observed by the pipeline. Trajectory error therefore reflects both
structural error in the discovered equation and parameter-estimation
error induced by noisy observations.

\paragraph{Form error.}
Form error isolates the quality of the discovered symbolic structure by
removing the effect of noisy parameter estimation. When the observations
are noisy, the fitted parameters of even a correct form are inexact, so a
large trajectory error does not necessarily indicate that the form is
wrong. We therefore discard $\hat{\theta}$ and re-estimate the parameters
of $f$ on noise-free data from the training regimes,
\[
\tilde\theta=\arg\min_{\theta>0}\ \mathrm{NMSE}(f,\theta;\mathcal{R}_{\mathrm{train}}),
\]
\[
E_{\mathrm{form}}=\mathrm{NMSE}(f,\tilde\theta;\cdot),
\]
and report $E_{\mathrm{form}}$ on the training and held-out regimes. The
minimisation uses the Stage~4 procedure with multiple starts. Without noise
the pipeline already fits noise-free data, so $E_{\mathrm{form}}$ coincides
with the trajectory error and is not reported.

\section{Results}

\subsection{Form error and Robustness to Observation Noise}
\label{sec:noise_results}

Table~\ref{tab:formerror} compares form error across the four
biological systems. On clean observations, P2C is not uniformly
superior: several baselines recover highly accurate forms when the
numerical evidence is informative. The difference becomes pronounced
as the observations are degraded. At $\sigma=0.05$, P2C obtains the
lowest form error on three of the four systems, and at
$\sigma=0.1$ it achieves the lowest error on all four systems.

This trend supports the central hypothesis behind P2C: explicit
mechanistic constraints provide relatively little advantage when the
trajectory itself strongly identifies the governing equation, but
become increasingly useful as numerical evidence becomes ambiguous. Errors of the unconstrained baselines increase sharply with noise while P2C remains substantially more stable.

\begin{table}
\centering
\caption{Form error comparison across ODE systems under different noise levels. Here D refers to the number of ODE systems, and lower values are better.}
\label{tab:noise_results}
\resizebox{\linewidth}{!}{
\begin{tabular}{c l c c c c}
\hline
$\sigma$ & Model & Dunster (4D) & Baker (2D) & Kumar (3D) & Reynolds (4D)\\
\hline

\multirow{3}{*}{0}
& LLM-SR & 2.412 & 0.056 & 0.34 & \textbf{0.089} \\
& DoLQ   & 1.76  & 0.043 & 0.0049 & 0.25 \\
& EDL    & \textbf{4.9e-11} & 0.0027 & \textbf{8.9e-4} & 0.28 \\
& P2C (Ours)  & 1.1e-9 & \textbf{0.0008} & 0.032 & 0.499 \\
\hline

\multirow{3}{*}{0.05}
& LLM-SR & 3.193 & 1.625 & 0.97 & 1.097 \\
& DoLQ   & 2.539 & 0.051 & 0.21 & 0.44 \\
& EDL    & 0.14 & 0.0045 & 1.19 & \textbf{0.35} \\
& P2C (Ours)   & \textbf{4.2e-11} & \textbf{0.0024} & \textbf{0.044} &  0.551\\
\hline

\multirow{3}{*}{0.1}
& LLM-SR & 10.939 & 5.498 & 9.3 & 2.014 \\
& DoLQ   & 3.098 & 1.604 & 0.23 & 360 \\
& EDL    & 1.605 & 0.013 & $\infty$ & 1.391 \\
& P2C (Ours)  & \textbf{1.4e-10} & \textbf{0.0047} & \textbf{0.024} & \textbf{1.095} \\
\hline
\end{tabular}
}
\label{tab:formerror}
\end{table}

\subsection{Importance of the Text2Constraint stage}\label{sec:memorisation}
 
\begin{figure}
    \centering
    \includegraphics[width=1\linewidth]{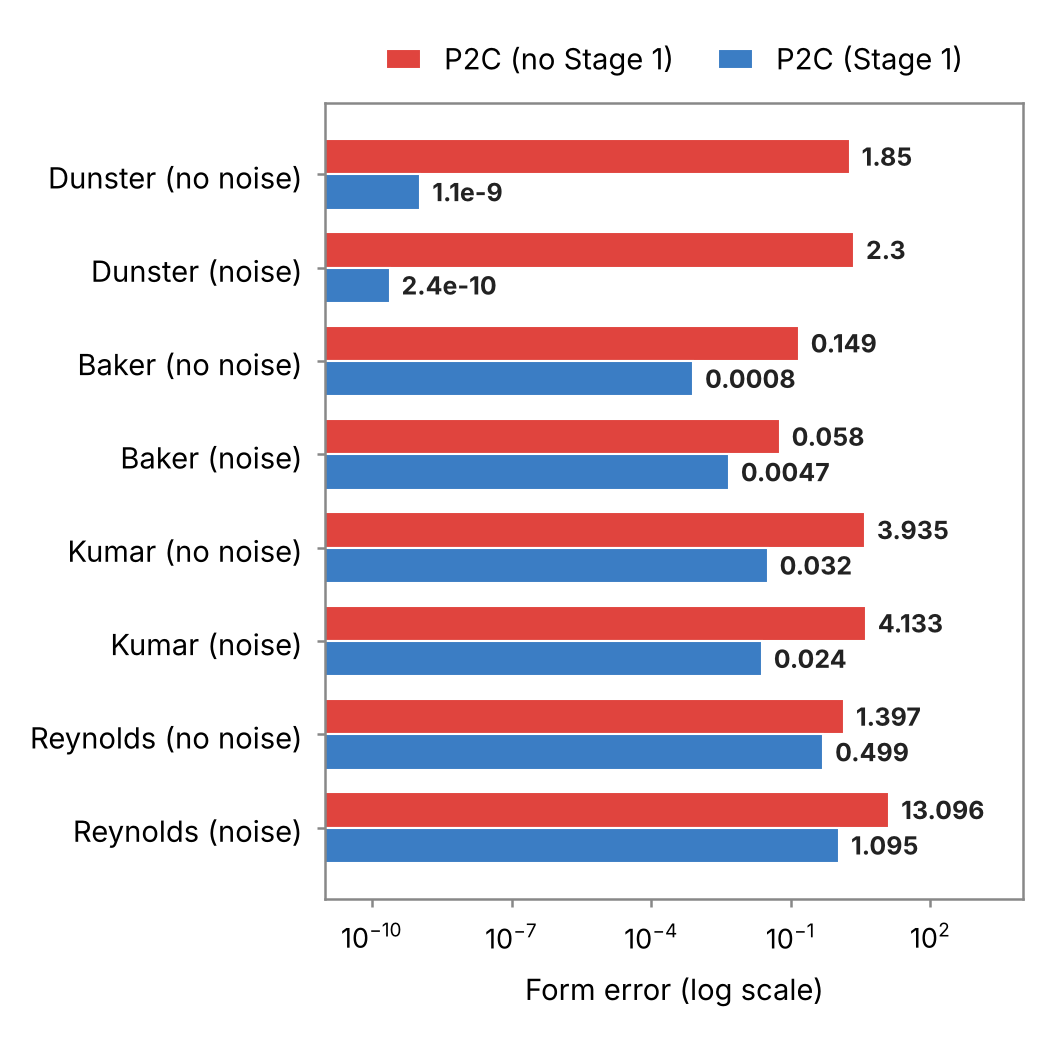}
    \caption{Form recovery error on the ODEs on clean data and noisy data ($\sigma = 0.1$) for P2C with and without the Text2Constraint stage (Stage 1). Lower values are better. P2C with stage 1 consistently performs better than the same framework without Stage 1.}
    \label{fig:P2CStage1}
    
\end{figure}

To isolate the contribution of explicitly grounding the biological
description into formal constraints, we compare the full P2C pipeline
with a variant in which Stage~1 is removed. All remaining components
of the discovery pipeline are kept unchanged. Figure \ref{fig:P2CStage1} shows that
Text2Constraint improves form recovery across all four systems, under
both clean and noisy observations.

The effect is particularly pronounced for Dunster: without
Text2Constraint, the form error is $1.85$ on clean data and $2.30$ at
$\sigma=0.1$, whereas the full pipeline reduces these errors to
$1.1\times10^{-9}$ and $2.4\times10^{-10}$, respectively. Substantial
improvements are also observed for Baker and Kumar. For Reynolds,
where the overall discovery problem remains more difficult,
Text2Constraint still reduces form error from $1.397$ to $0.499$ on
clean data and from $13.096$ to $1.095$ under noise.

These results indicate that the gains of P2C do not arise solely from
the downstream LLM search and parameter fitting. Converting the prose into explicit, verifiable constraints filters candidate structures that violate the described biological mechanism. The constraints do not uniquely determine an equation; rather, they remove mechanistically inconsistent candidates while numerical evidence distinguishes among the remaining structures.

% \begin{table}
% \centering\small
% \setlength{\tabcolsep}{4pt}
% \begin{tabular}{@{}lccc@{}}
% \toprule
% System & P2C (no Stage 1) & P2C (Stage 1)\\
% \midrule
% Dunster (no noise)  & 1.85 & $1.1\times10^{-9}$ \\
% Dunster (noise)  & 2.3 & $2.4\times10^{-10}$\\
% Baker (no noise) & 0.149 & 0.0008\\
% Baker (noise) & 0.058 & 0.0047\\
% Kumar (no noise)  & 3.935 & 0.032\\
% Kumar (noise)  & 4.133 & 0.024\\
% Reynolds (no noise)  & 1.397 & 0.499\\
% Reynolds (noise)  & 13.096 & 1.095\\
% \bottomrule
% \end{tabular}
% \caption{Form recovery error on the ODEs on clean data and noisy data ($\sigma = 0.1$) for P2C with and without the Text2Constraint stage.}
% \label{tab:P2CStage1}
% \end{table}

% \begin{table}
% \centering\small
% \setlength{\tabcolsep}{4pt}
% \begin{tabular}{@{}lccc@{}}
% \toprule
% System & No text & P2C \\
% \midrule
% Dunster (no noise)  & 0 & 100 \\
% Dunster (noise)  & 0 & 100 \\
% Baker (no noise) & 0 & 67\\
% Baker (noise) & 0 & 25\\
% \bottomrule
% \end{tabular}
% \caption{Form recovery rate (\%) on the ODEs and their variants at $\sigma=0.1$.}
% \label{tab:memor}
% \end{table}
 
% ---------------------------------------------------------------------
% \subsection{Search cost}\label{sec:cost}
 
% Symbolic verification runs before fitting, so rejected candidates cost no
% fit. Table~\ref{tab:cost} reports the cost per run.
 
% \TODO{Fill:} Stage~3 rejects \XX\% of proposed candidates before fitting. The
% most frequently violated constraint types are \XX.
 
% ---------------------------------------------------------------------
\subsection{Trajectory error}\label{sec:errors}

\begin{figure}
    \centering
    \includegraphics[width=1\linewidth]{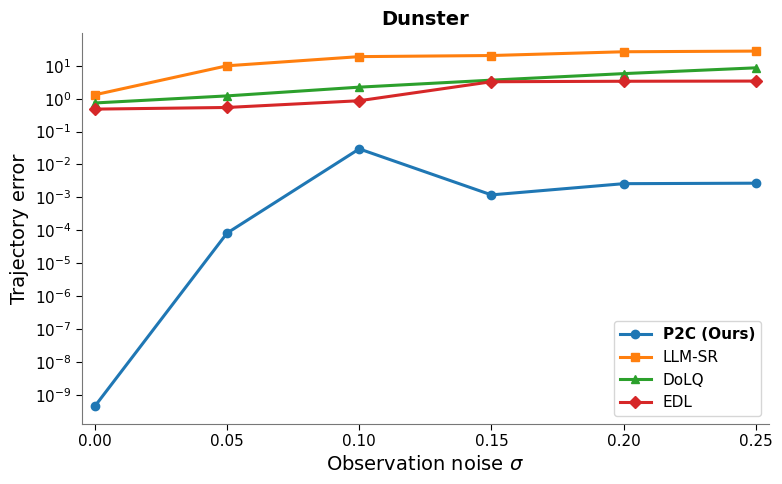}
    \includegraphics[width=1\linewidth]{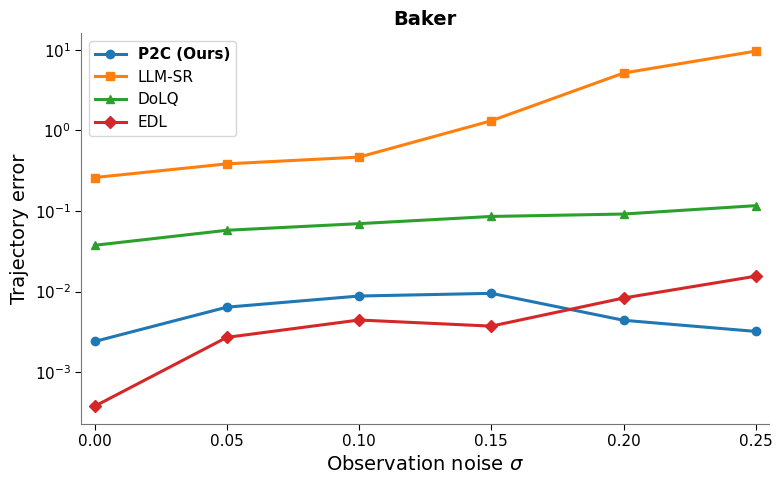}
    \includegraphics[width=1\linewidth]{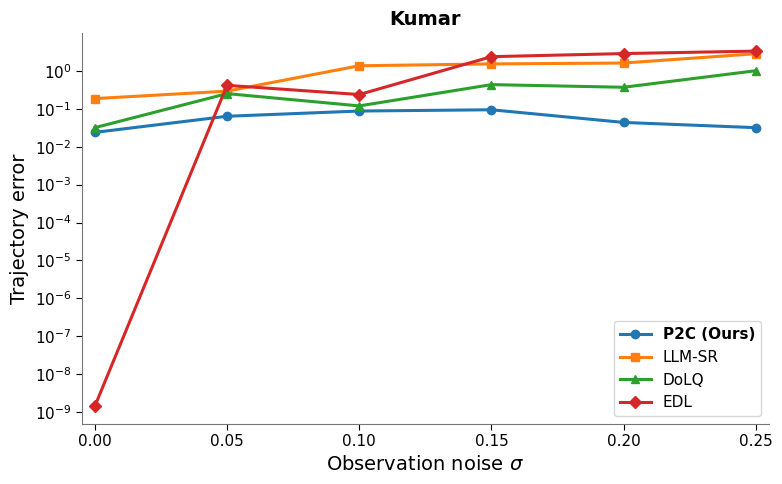}
    \includegraphics[width=1\linewidth]{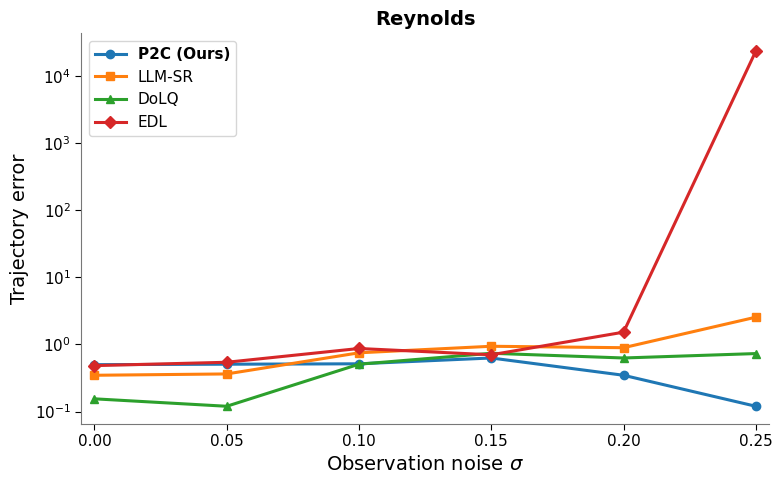}
    \caption{Trajectory error as a function of observation noise for P2C and baselines. While initially P2C performs comparable to some other baselines, it consistently does better than baseline models as noise increases. Lower values are better.}
    \label{fig:trajectory_error}   
\end{figure}

Figure \ref{fig:trajectory_error} shows the trajectory error as the observation noise increases. P2C achieves substantially lower error than the LLM-based baselines over the evaluated noise range. As noise
increases, the trajectory errors of LLM-SR, DoLQ, and EDL rise
considerably, while P2C remains comparatively stable. The gap is
especially notable once the observations become noisy, consistent
with the form-error results in Section~5.1.

The difference between the two metrics is also informative. A model
may recover an appropriate symbolic form yet exhibit non-negligible
trajectory error because its parameters were estimated from corrupted
observations. Conversely, the consistently lower trajectory error of
P2C indicates that its advantage is not limited to recovering
plausible symbolic structure: the constrained search also yields
models that can be fitted more reliably and reproduce the underlying
dynamics more accurately.

\section{Application in Therapeutic Discovery}

While the P2C framework demonstrates robust equation recovery on standard benchmark systems, its primary clinical value lies in modeling therapeutic interventions where observational data is sparse, noisy, heterogeneous, and expensive to obtain \cite{Alber2019Integrating}. One such application is predicting optimal parameters for peripheral focused ultrasound (pFUS) therapy in treating inflammatory arthritis \cite{Zachs2019Noninvasive, Cotero2019Noninvasive}. The available arthritis data \cite{Zachs2019Noninvasive} is limited (50 samples). Under these data-constrained conditions, multiple equation forms can fit the same observations equally well during the training phase, yet encode vastly different physical mechanisms when extrapolated beyond the observed regime \cite{Gutenkunst2007Sloppy, Raue2009Identifiability, Kronberger2022ShapeConstrained}. Hence standard symbolic regression methods struggle to identify equations that both fit into the data as well as account for the true underlying biological mechanism.

% When tasked with discovering the governing ordinary differential equations (ODEs) for ultrasound treatment using only numerical data, existing LLM-based equation discovery methods default to naive polynomial or exponential fits. Because the empirical data captures initial improvement as ultrasound duration increases, data-only models propose equations with a monotonically increasing relationship between treatment time and healing trajectory. However, this mathematical extrapolation is biologically impossible; applying focused ultrasound continuously for several hours would induce severe thermal toxicity, cause tissue burns, and potentially damage internal organs.

To address this modeling gap, P2C incorporates mechanistic knowledge expressed in scientific text regarding the Cholinergic Anti-Inflammatory Pathway (CAP) \cite{Tracey2002InflammatoryReflex, Kelly2022InflammatoryReflex, RosasBallina2011Acetylcholine}. We claim that on such a real-world system, P2C discovers bounded, asymptotic ODEs that more accurately reflect the biological realities of the ultrasound-based therapy. 

\subsection{Experiment: Duration of splenic ultrasound}
\label{sec:app-exp1}
 
\paragraph{Data and model.}
We use the dose-response experiment of \citet{Zachs2019Noninvasive}: mice receive
arthritogenic K/BxN serum on day~0 and daily ultrasound of the spleen
(1~MHz, 350~kPa) for $u \in \{0, 2, 6, 12, 20\}$ minutes per day; ankle
swelling $A$ (increase of ankle thickness, mm) is measured on days 0-7.
We use the median of each group (16, 16, 6, 6 and 6 mice). The task is to find one equation $\dot A = f(A, u)$, with $u$ a known input that differs between groups.
The 20-minute group is held out from all methods; the other four groups
are the training data. 
 
\paragraph{Description.}
The LLM writes the description from seven short notes that summarise
published work on the CAP pathway \cite{Tracey2002InflammatoryReflex,RosasBallina2011Acetylcholine,Cotero2019Noninvasive,Zachs2019Noninvasive, Koopman2016Vagus, Levine2014Neurostimulation, Wang2003Nicotinic}: the LLM first traces the pathway from the stimulus to the measured variable and then writes sentences about the model variables, without equations, numbers or results of the experiment (prompt in Appendix~\ref{app:prose}). All methods receive the same description.
 
\paragraph{Methods and metrics.}
We compare P2C with the no-text variant and with LLM-SR, DoLQ and EDL,
all with GPT-6-Luna and three seeds. We report the NMSE on the held-out group with the parameters returned by each method and the number of reference constraints that the returned equation violates (Appendix~\ref{app:prose}).
 
% \begin{table}[t]
% \centering
% \small
% \begin{tabular}{@{}lccccc@{}}
% \toprule
% Method  & NMSE & RMSE & Violations \\
% \midrule
% LLM-SR     & 1.97 & 0.044 & 3\\
% DoLQ       & 5.76 & 0.103 & 2 \\
% EDL        & 1.97 & 0.105 & 2 \\
% P2C        & \textbf{0.22} & \textbf{0.035} & 0 \\
% \bottomrule
% \end{tabular}
% \caption{Ankle swelling NMSE and the number of constraints violated on the held-out validation group of the Arthritis data \cite{Zachs2019Noninvasive}}
% \label{tab:app-exp1}
% \end{table}
 
\paragraph{Results.}

\begin{figure}
    \centering
    \includegraphics[width=1\linewidth]{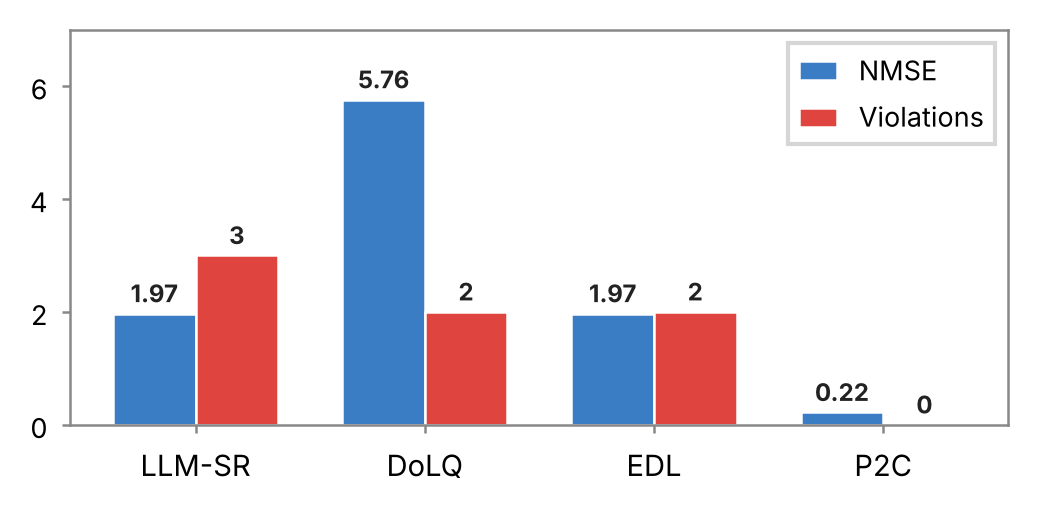}
    \caption{Ankle swelling NMSE and the number of constraints violated on the held-out validation group of the Arthritis data \cite{Zachs2019Noninvasive}. Lower is better for both metrics.} 
    \label{fig:app-exp1}
\end{figure}

Figure \ref{fig:app-exp1} shows that P2C produces equations that are more consistent with the mechanism described in the literature, in addition to achieving lower held-out prediction error. The reference description
induces four mechanistic constraints: swelling must have a positive
source after serum transfer, remain non-negative, decrease in rate as
ultrasound duration increases, and exhibit a bounded effect as the
duration becomes large. P2C has a median of zero violated constraints
across the three runs, compared with four for LLM-SR and two for both
DoLQ and EDL. In two of the three P2C runs, all four constraints are
satisfied, yielding $\dot A = \frac{0.153}{1+0.31u}$, which predicts a positive swelling rate that decreases monotonically with ultrasound duration and approaches a finite limit. The remaining P2C run violates one reference constraint. In contrast, several baseline equations violate multiple properties stated by the pathway description, despite being fitted to the same observations. Thus, the advantage of P2C in this application is not only lower held-out NMSE but also greater agreement between the discovered equation and the mechanistic knowledge expressed in scientific text.

\section{Related Work}
\label{sec:related}

\paragraph{Equation discovery.}
Symbolic regression searches for mathematical expressions that explain observed data
\citep{schmidt2009distilling,cranmer2023pysr}, while SINDy selects terms from a predefined
library \citep{brunton2016sindy} and ODEFormer predicts ODEs directly from trajectories
\citep{dascoli2024odeformer}. Because numerical differentiation can amplify observation
noise, weak-form approaches instead transfer derivatives to smooth test functions
\citep{Messenger2021WeakSINDy}. Sparse and noisy observations can nevertheless leave models
weakly constrained or non-identifiable
\citep{Raue2009Identifiability,Gutenkunst2007Sloppy}.

\paragraph{LLMs for equation discovery.}
Recent methods use LLMs to propose or refine equation structures, including LLM-SR
\citep{shojaee2025llmsr}, learned concept libraries \citep{grayeli2024lasr},
in-context refinement \citep{merler2024icsr}, evolutionary search
\cite{10.1063/5.0224297}, agentic code generation \citep{holt2024datadriven}, and qualitative
candidate assessment \citep{song2026discovering}. Other work incorporates LLM-derived
plausibility into the regression objective \citep{Taskin2026Knowledge}. These methods typically
use scientific knowledge as free-text guidance or a score, rather than explicitly
extracting and verifying the mechanistic information supplied by the text.

\paragraph{Text-derived constraints.}
Prior work incorporates expert-specified constraints into symbolic regression through
shape constraints \citep{Kronberger2022ShapeConstrained}, formal verification
\citep{bladek2019formal}, prior-knowledge objectives \citep{kubalik2020prior},
and side information \citep{machado2023sparseidentificationnonlineardynamics}.
Separately, systems such as INDRA extract executable biological models from text
\citep{gyori2017}, while neuro-symbolic approaches translate language into formal
representations for reasoning \citep{pan2023logiclm,olausson2023linc}.
Concurrent MEDA retrieves background knowledge and generates constraints for biological
ODE discovery \citep{krongauz2026meda}.

P2C connects these directions by translating a fixed mechanistic description into
typed, text-grounded constraints that are checked against candidate ODEs before
parameter fitting. It also supports non-polynomial structures such as saturating
functions.

\section{Conclusion}

We introduced P2C, an agentic framework that translates biological
prose into explicit, verifiable constraints for ODE discovery before
numerical fitting. Across four published models of inflammation and
immune regulation, P2C becomes increasingly advantageous as
observations are degraded, while ablations show that the
Text2Constraint stage contributes substantially to equation-form
recovery. We further demonstrate the framework on measured
dose--response data from splenic ultrasound treatment of inflammatory
arthritis, where mechanistic constraints help favor bounded,
biologically plausible equations. Overall, our results suggest that
scientific text can provide useful structural information when
numerical observations alone are insufficient to identify the
governing dynamics. Future work will extend P2C to larger biological
systems and richer forms of mechanistic knowledge.

\section{Limitations}

% Our evaluation is limited to four ODE systems from inflammatory and
% immune biology, with mechanistic descriptions manually derived from
% the corresponding source models. The current constraint language also
% captures only a restricted set of mechanistic properties, so the
% results do not establish generalization to substantially larger
% systems or other scientific domains.

% P2C further assumes that the supplied mechanistic description is a
% reasonably coherent summary of the relevant scientific knowledge. We
% do not evaluate robustness to conflicting literature, ambiguous or
% unsupported statements, or mechanisms requiring substantial
% domain-specific interpretation. The grounding checks verify
% consistency with the provided description, not the scientific
% correctness of that description.

Our controlled benchmark contains four ODE systems from inflammatory
and immune biology and is intended to test whether text-derived
constraints improve equation discovery under known mechanisms. It
does not establish broad generalization across the scientific
literature, substantially larger dynamical systems, or other domains.
The current constraint schema also captures only a restricted subset
of mechanistic properties and filters, rather than uniquely determines,
the admissible equation structures.

P2C assumes that the supplied mechanistic description is a reasonably
coherent summary of the relevant scientific knowledge. We do not
evaluate robustness to conflicting literature, ambiguous or
unsupported statements, or mechanisms requiring substantial
domain-specific interpretation. The grounding checks verify
consistency with the provided description, not the scientific
correctness of that description.

\section{Ethical Consideration}

% To the best of our knowledge, we did not violate any ethical code while conducting the research work described in this paper. The paper is a new method of doing equation discovery, motivating it's importance especially in biology and medicine. However, the method does not guarantee new universal therapy.  No new human
% or animal experiments were conducted as part of this work. The research uses existing published articles and datasets for our experiments. 

To the best of our knowledge, this work was conducted in accordance
with applicable ethical standards. We introduce a new method for
equation discovery, motivated in part by applications in biology and
medicine. The discovered equations are intended as exploratory
mechanistic models and do not constitute clinical recommendations,
treatment-selection tools, or guarantees of a new or universally
effective therapy.

Because the method relies on mechanistic descriptions derived from
scientific text, incorrect, incomplete, or unsupported descriptions
could lead to misleading constraints and consequently inappropriate
model forms. Any biomedical use would therefore require independent
experimental validation and review by domain experts before the
results could inform practice.

No new human or animal experiments were conducted as part of this
work. All experiments use previously published articles and existing
datasets.

We release prompts, prose descriptions and implementation details in the Appendix C. Third-party datasets and source materials are not redistributed and should be obtained and cited from their original sources. 

An AI assistant was used during manuscript preparation for language
editing, restructuring of selected passages, and assistance with
drafting and revising explanatory text. It was also used for limited
support in literature-search planning and code-related discussion.
All technical claims, citations, experimental results, and final
manuscript content were reviewed and verified by the authors.

\section{Acknowledgements}

We would like to thank the Indiana Clinical and Translational Sciences Institute (CTSI), through its Project Development Team (PDT) program, for funding support for the project titled “Novel Ultrasound Therapy and AI-Guided Modeling for Burn Treatment.” We also thank Naman Saxena and Shaiv Mehra for their contributions to the early-stage ideation and conceptual development of this work.

% Custom bibliography entries only
\bibliography{custom}

\appendix

\section{Dataset description}
\label{sec:app-data}

\paragraph{Mechanistic models of inflammation.}
Our benchmark systems are reduced ODE models that were built by hand from
qualitative knowledge of the kind that P2C takes as input.
\citet{Kumar2004Dynamics} describe acute inflammation with three variables, a pathogen
and an early and a late pro-inflammatory mediator; the model reproduces the
healthy outcome and several negative outcomes, depending on the initial
conditions and the parameters.
\citet{Reynolds2006Reduced} derive a four-variable model of the response to infection
(pathogen, activated phagocytes, tissue damage and an anti-inflammatory
mediator) from the direct interactions of these effectors, and use it to study
how anti-inflammation promotes resolution.
\citet{Dunster2014Resolution} model the resolution of inflammation through the
interaction of neutrophils and macrophages (our version has four variables:
neutrophils, apoptotic neutrophils, macrophages and a pro-inflammatory
mediator); the model has a healthy and an unhealthy outcome, and the rates of
macrophage phagocytosis and neutrophil apoptosis decide between them.
\citet{Baker2013CytokineInflammation} reduce cytokine regulation in rheumatoid arthritis to two
variables, the pro-inflammatory and the anti-inflammatory cytokines, and
obtain bistability and oscillations. 

% \begin{table*}[t]
% \centering
% \small
% \begin{tabular}{@{}llp{3.5cm}p{3.0cm}p{3.4cm}@{}}
% \toprule
% System & $d$ & State variables \\
% \midrule
% Kumar \citep{Kumar2004Dynamics} & 3 &
%   pathogen; early and late pro-inflammatory mediators \\
% Reynolds \citep{Reynolds2006Reduced} & 4 &
%   pathogen; activated phagocytes; tissue damage; anti-inflammatory mediator 
% Dunster \citep{Dunster2014Resolution} & 4 &
%   neutrophils; apoptotic neutrophils; macrophages; pro-inflammatory mediator &
%   rate at which macrophages remove apoptotic neutrophils &
%   Hill term for the release of the mediator \\
% Baker \citep{Baker2013CytokineInflammation} & 2 &
%   pro-inflammatory and anti-inflammatory cytokines &
%   clearance rate of the pro-inflammatory cytokines &
%   Hill terms for production; inhibition by the anti-inflammatory cytokines \\
% \bottomrule
% \end{tabular}
% \caption{Benchmark systems. $d$: number of state variables. The known input
% differs between regimes and is given to every method.}
% \label{tab:systems}
% \end{table*}

\paragraph{Simulation protocol.}
For every system and regime we integrate the equations below from the
listed initial state over $[0, T]$ with LSODA \citep{petzold1983lsoda}
(relative tolerance $10^{-10}$, absolute tolerance $10^{-12}$) and record
201 equally spaced time points, including both end points. Values below
zero are set to zero. Each system has one known input that differs between
regimes; it is given to every method. Three regimes are used for training
and one, \emph{held out}, is used only for evaluation. For noisy data,
Gaussian noise with standard deviation $\sigma$ times the root mean square
of each state over the regime is added independently to every point, and
values below zero are again set to zero. P2C fits on the points with even
index and scores candidates on the points with odd index; the baselines
receive all points of the three training regimes. Parameters are fixed to
the values below in all regimes; only the input and the initial state
change.
 
\paragraph{Reynolds.}
Four states: pathogen $P$, activated phagocytes $N^{*}$, tissue damage $D$
and anti-inflammatory mediator $C_A$ \citep[Eqs.~9--12 and Table~1]{Reynolds2006Reduced}.
The input is the growth rate of the pathogen, $k_{pg}$. With the
inhibition function $f(V) = V/\bigl(1 + (C_A/c_\infty)^2\bigr)$ and
$R = f(k_{nn}N^{*} + k_{np}P + k_{nd}D)$,

\begin{align*}
\dot P   &= k_{pg}P\Bigl(1 - \frac{P}{p_\infty}\Bigr)
            - \frac{k_{pm}s_m P}{\mu_m + k_{mp}P} - k_{pn}f(N^{*})\,P,\\
\dot N^{*} &= \frac{s_{nr}R}{\mu_{nr} + R} - \mu_n N^{*},\\
\dot D   &= \frac{k_{dn}f(N^{*})^6}{x_{dn}^6 + f(N^{*})^6} - \mu_d D,\\
\dot C_A &= s_c + \frac{k_{cn}\,f(N^{*} + k_{cnd}D)}{1 + f(N^{*} + k_{cnd}D)} - \mu_c C_A .
\end{align*}
Ground-truth parameters: $k_{pm}=0.6$, $k_{mp}=0.01$, $s_m=0.005$, $\mu_m=0.002$,
$p_\infty=20$, $k_{pn}=1.8$, $k_{np}=0.1$, $k_{nn}=0.01$, $k_{nd}=0.02$,
$s_{nr}=0.08$, $\mu_{nr}=0.12$, $\mu_n=0.05$, $k_{dn}=0.35$,
$x_{dn}=0.06$, $\mu_d=0.02$, $c_\infty=0.28$, $s_c=0.0125$,
$k_{cn}=0.04$, $k_{cnd}=48$, $\mu_c=0.1$.
 
\paragraph{Kumar.}
Three states: pathogen $p$, early pro-inflammatory mediators $m$ and late
pro-inflammatory mediators $l$ \citep{Kumar2004Dynamics}; $p$ and $m$ are fractions
of their maximum levels. The input is the rate $k_{pm}$ at which early
mediators destroy the pathogen. With $f(m) = 1 + \tanh\bigl((m-\theta)/w\bigr)$,
\begin{align*}
\dot p &= k_p\,p(1-p) - k_{pm}\,m\,p,\\
\dot m &= (k_{mp}\,p + l)\,m(1-m) - m,\\
\dot l &= k_{lm}\,f(m) - k_l\,l .
\end{align*}
Ground-truth parameters:: $k_p=3$, $k_{mp}=25$, $k_{lm}=15$, $k_l=1$, $\theta=1$, $w=0.5$. Every regime starts from the resting level of the late mediators,
$l_0 = k_{lm}\bigl(1+\tanh(-\theta/w)\bigr)/k_l \approx 0.5396$.
 
\paragraph{Dunster.}
Four states: neutrophils $n$, pro-inflammatory mediator $c$, apoptotic
neutrophils $a$ and macrophages $m$ \citep{Dunster2014Resolution}. We use the
model without the time-dependent external stimulus of the original, so inflammation is driven only by the initial mediator level.
The input is the rate $\phi$ at which macrophages remove apoptotic
neutrophils:
\begin{align*}
\dot n &= c - \nu n,\\
\dot c &= \frac{\gamma_a a^2}{\beta_a^2 + a^2} - c,\\
\dot a &= \nu n - \gamma_a a - \phi\,a\,m,\\
\dot m &= c - \gamma_m m .
\end{align*}
Ground-truth parameters:: $\nu=0.1$, $\gamma_a=1$, $\beta_a=0.1$, $\gamma_m=0.01$.
 
\paragraph{Baker.}
Two states: anti-inflammatory cytokines $a$ and pro-inflammatory cytokines
$p$ \citep{Baker2013CytokineInflammation}. The input is the clearance rate $k_5$ of the
pro-inflammatory cytokines:
\begin{align*}
\dot a &= \frac{c_4 p^2}{c_3^2 + p^2} - a,\\
\dot p &= \frac{c_1 + c_2\,p^2/(1+p^2)}{1 + a^2} - k_5\,p .
\end{align*}
Ground-truth parameters:: $c_1=0.025$, $c_2=1$, $c_3=0.5$, $c_4=3.5$.
 
\begin{table*}[t]
\centering
\small
\begin{tabular}{@{}lllll l@{}}
\toprule
System & Regime & Input & Initial state & $T$ & Use \\
\midrule
Reynolds & health   & $k_{pg}=0.30$ & $(P,N^{*},D,C_A)=(0.5,\,0,\,0,\,0.125)$ & 100 & training \\
         & aseptic  & $k_{pg}=0.30$ & $(1.5,\,0,\,0,\,0.125)$ & 100 & training \\
         & septic   & $k_{pg}=0.60$ & $(1.0,\,0,\,0,\,0.125)$ & 100 & training \\
         & held out & $k_{pg}=0.45$ & $(1.2,\,0,\,0,\,0.125)$ & 100 & evaluation \\
\midrule
Kumar    & low dose & $k_{pm}=30$ & $(p,m,l)=(0.01,\,0.05,\,l_0)$ & 20 & training \\
         & aseptic  & $k_{pm}=30$ & $(0.20,\,0.05,\,l_0)$ & 20 & training \\
         & septic   & $k_{pm}=3$  & $(0.01,\,0.05,\,l_0)$ & 20 & training \\
         & held out & $k_{pm}=10$ & $(0.05,\,0.05,\,l_0)$ & 20 & evaluation \\
\midrule
Dunster  & chronic  & $\phi=0.01$ & $(n,c,a,m)=(0,\,0.2,\,0,\,0)$ & 100 & training \\
         & slow     & $\phi=0.10$ & $(0,\,1.0,\,0,\,0)$ & 100 & training \\
         & resolved & $\phi=1.00$ & $(0,\,1.0,\,0,\,0)$ & 100 & training \\
         & held out & $\phi=0.30$ & $(0,\,0.5,\,0,\,0)$ & 100 & evaluation \\
\midrule
Baker    & fast clearance   & $k_5=1.25$ & $(a,p)=(0,\,2.00)$ & 30 & training \\
         & slow clearance   & $k_5=0.30$ & $(0,\,0.05)$ & 30 & training \\
         & medium clearance & $k_5=0.60$ & $(0,\,1.00)$ & 30 & training \\
         & held out         & $k_5=0.40$ & $(0,\,0.50)$ & 30 & evaluation \\
\bottomrule
\end{tabular}
\caption{Regimes of the four benchmark systems. \emph{Initial state}: in the order of the first row of each system ($t=0$); $T$: simulation horizon, in the time units of the original model; $l_0 \approx 0.5396$ is the resting level of the late mediators in Kumar. Each regime has 201 equally spaced time points on $[0,T]$. The held-out
regime is never seen by any method and is used only for evaluation.}
\label{tab:regimes}
\end{table*}
 
\subsection{Descriptions and gold constraints}
\label{app:gold}
 
Tables~\ref{tab:gold-reynolds}--\ref{tab:gold-baker} give the full
description of each benchmark system, sentence by sentence, in the order in
which it is given to every method. Next to each sentence are the ground truth (gold)
constraints that it states, written by hand, and the predicate that the
checker of Stage~3 tests for each (Table~\ref{tab:schema}). Here $f_x$
denotes the right-hand side of the equation for state $x$, and all
predicates are decided for positive states, inputs and parameters. A
sentence without a constraint states a mechanism that the schema cannot
express, such as logistic growth or self-amplification; these sentences
are kept so that the description reads as a whole, and their content is
left to the data. The
descriptions contain no equation, number or name of a function.

\begin{table}[t]
\centering
\footnotesize
\setlength{\tabcolsep}{2.5pt}
\resizebox{\columnwidth}{!}{%
\begin{tabular}{@{}lccc cccccc c@{}}
\toprule
System & $d$ & Sent. & Empty & \textsc{Sign} & \textsc{Abs} & \textsc{Sat} & \textsc{Int} & \textsc{Src} & \textsc{Pos} & Total \\
\midrule
\textsc{Reynolds} & 4 & 20 & 3 & 11 & 3 & 4 & 1 & 1 & 4 & 24 \\
\textsc{Kumar} & 3 & 13 & 2 & 5 & 2 & 1 & 3 & 1 & 3 & 15 \\
\textsc{Dunster} & 4 & 13 & 0 & 9 & 3 & 1 & 1 & 0 & 4 & 18 \\
\textsc{Baker} & 2 & 9 & 2 & 3 & 0 & 1 & 1 & 1 & 2 & 8 \\
\midrule
Total & & 55 & 7 & 28 & 8 & 7 & 6 & 3 & 13 & 65 \\
\bottomrule
\end{tabular}}
\caption{Number of gold constraints by system and type. $d$: states; Sent stands for sentences in the prose description. Empty stands for sentences that state no
constraint. Complete prose descriptions are in Tables~\ref{tab:gold-reynolds}--\ref{tab:gold-baker}.}
\label{tab:gold-summary}
\end{table}

\begin{table*}[p]
\centering
\footnotesize
\renewcommand{\arraystretch}{1.15}
\begin{tabular}{@{}r p{0.52\textwidth} p{0.17\textwidth} p{0.21\textwidth}@{}}
\toprule
\# & Sentence of the description & Gold constraint & Predicate \\
\midrule
1 & The pathogen multiplies on its own, so its amount grows faster when more of it is present, but growth slows as the pathogen approaches the maximum load that the host can carry. & -- & -- \\
2 & A fast background defence that is present before any insult removes some pathogen, but this defence is limited and is overwhelmed when the pathogen is abundant. & -- & -- \\
3 & Activated phagocytes consume the pathogen on contact, so the pathogen falls faster when more activated phagocytes are present. & \textsc{Sign}$(P,N^{*},-)$ \newline \textsc{Inter}$(P;P,N^{*})$ & $\partial f_{P}/\partial N^{*} < 0$ \newline a term of $f_{P}$ has $P$ and $N^{*}$ \\
4 & Tissue damage does not itself act on the pathogen. & \textsc{Absent}$(P,D)$ & $\partial f_{P}/\partial D \equiv 0$ \\
5 & The anti-inflammatory mediator reduces the ability of activated phagocytes to attack the pathogen. & \textsc{Sign}$(P,C_A,+)$ & $\partial f_{P}/\partial C_A > 0$ \\
6 & Resting phagocytes are recruited and activated by the pathogen and by tissue damage, so more pathogen or more damage makes activated phagocytes rise faster. & \textsc{Sign}$(N^{*},P,+)$ \newline \textsc{Sign}$(N^{*},D,+)$ & $\partial f_{N^{*}}/\partial P > 0$ \newline $\partial f_{N^{*}}/\partial D > 0$ \\
7 & Activated phagocytes also recruit further phagocytes through the signals that they release, and they die off naturally over time. & -- & -- \\
8 & Recruitment cannot grow without bound, because the pool of resting phagocytes available to be activated is finite. & \textsc{Sat}$(N^{*},P)$ \newline \textsc{Sat}$(N^{*},D)$ & $\lim_{P\to\infty} f_{N^{*}}$ finite \newline $\lim_{D\to\infty} f_{N^{*}}$ finite \\
9 & The anti-inflammatory mediator suppresses the activation of phagocytes, so more mediator makes activated phagocytes rise more slowly. & \textsc{Sign}$(N^{*},C_A,-)$ & $\partial f_{N^{*}}/\partial C_A < 0$ \\
10 & Activated phagocytes inflict collateral damage on healthy tissue, so more activated phagocytes make damage rise faster. & \textsc{Sign}$(D,N^{*},+)$ & $\partial f_{D}/\partial N^{*} > 0$ \\
11 & Once phagocytes are abundant, further accumulation adds little additional damage. & \textsc{Sat}$(D,N^{*})$ & $\lim_{N^{*}\to\infty} f_{D}$ finite \\
12 & The pathogen does not damage tissue directly; damage arises only through the phagocytic response that it provokes. & \textsc{Absent}$(D,P)$ & $\partial f_{D}/\partial P \equiv 0$ \\
13 & The anti-inflammatory mediator also limits the damage that activated phagocytes inflict. & \textsc{Sign}$(D,C_A,-)$ & $\partial f_{D}/\partial C_A < 0$ \\
14 & Damaged tissue is repaired and regenerated over time, so damage falls faster when more of it is present. & \textsc{Sign}$(D,D,-)$ & $\partial f_{D}/\partial D < 0$ \\
15 & A healthy organism maintains a background production of the anti-inflammatory mediator even with no insult present. & \textsc{Source}$(C_A)$ & $f_{C_A}(\mathbf{0}) > 0$ \\
16 & Both activated phagocytes and damaged tissue drive production of the anti-inflammatory mediator. & \textsc{Sign}$(C_A,N^{*},+)$ \newline \textsc{Sign}$(C_A,D,+)$ & $\partial f_{C_A}/\partial N^{*} > 0$ \newline $\partial f_{C_A}/\partial D > 0$ \\
17 & The pathogen does not induce the anti-inflammatory mediator directly. & \textsc{Absent}$(C_A,P)$ & $\partial f_{C_A}/\partial P \equiv 0$ \\
18 & Production of the mediator reaches a ceiling at high levels of inflammation, and the mediator holds back its own production as its level rises. & \textsc{Sat}$(C_A,N^{*})$ & $\lim_{N^{*}\to\infty} f_{C_A}$ finite \\
19 & The mediator is cleared from the system at a steady rate. & \textsc{Sign}$(C_A,C_A,-)$ & $\partial f_{C_A}/\partial C_A < 0$ \\
20 & All four quantities are concentrations and remain non-negative. & \textsc{Pos}$(P)$ \newline \textsc{Pos}$(N^{*})$ \newline \textsc{Pos}$(D)$ \newline \textsc{Pos}$(C_A)$ & $P=0 \Rightarrow f_{P} \ge 0$ \newline $N^{*}=0 \Rightarrow f_{N^{*}} \ge 0$ \newline $D=0 \Rightarrow f_{D} \ge 0$ \newline $C_A=0 \Rightarrow f_{C_A} \ge 0$ \\
\bottomrule
\end{tabular}
\caption{\textsc{Reynolds} \citep{Reynolds2006Reduced}: description and gold constraints
(20 sentences, 24 constraints; 3 sentences state no constraint).
States: $P$ (pathogen), $N^{*}$ (activated phagocytes), $D$ (tissue damage), $C_A$ (anti-inflammatory mediator).}
\label{tab:gold-reynolds}
\end{table*}
 
\begin{table*}[p]
\centering
\footnotesize
\renewcommand{\arraystretch}{1.15}
\begin{tabular}{@{}r p{0.52\textwidth} p{0.17\textwidth} p{0.21\textwidth}@{}}
\toprule
\# & Sentence of the description & Gold constraint & Predicate \\
\midrule
1 & The pathogen multiplies on its own, so its amount grows faster when more of it is present, but growth slows as the pathogen approaches the maximum load that the host can carry. & -- & -- \\
2 & Early pro-inflammatory mediators destroy the pathogen when they encounter it, so the pathogen falls faster when more early mediators are present. & \textsc{Sign}$(p,m,-)$ \newline \textsc{Inter}$(p;p,m)$ & $\partial f_{p}/\partial m < 0$ \newline a term of $f_{p}$ has $p$ and $m$ \\
3 & Late pro-inflammatory mediators do not act on the pathogen directly; they affect it only through the early mediators. & \textsc{Absent}$(p,l)$ & $\partial f_{p}/\partial l \equiv 0$ \\
4 & The pathogen activates the early mediators, so more pathogen makes the early response rise faster. & \textsc{Sign}$(m,p,+)$ & $\partial f_{m}/\partial p > 0$ \\
5 & The late mediators activate the early mediators further, which lets inflammation sustain itself after it has started. & \textsc{Sign}$(m,l,+)$ & $\partial f_{m}/\partial l > 0$ \\
6 & Both kinds of activation act on the early response that is already under way, and both slow as the early response approaches its maximum level. & \textsc{Inter}$(m;p,m)$ \newline \textsc{Inter}$(m;l,m)$ & a term of $f_{m}$ has $p$ and $m$ \newline a term of $f_{m}$ has $l$ and $m$ \\
7 & Early mediators are also lost at a steady rate, so the early response fades when nothing activates it. & -- & -- \\
8 & Early mediators drive the production of late mediators, so a stronger early response makes late mediators rise faster. & \textsc{Sign}$(l,m,+)$ & $\partial f_{l}/\partial m > 0$ \\
9 & Late mediators are produced in quantity only when the early response is strong, and their production cannot rise beyond a ceiling however strong the early response becomes. & \textsc{Sat}$(l,m)$ & $\lim_{m\to\infty} f_{l}$ finite \\
10 & A small background production of late mediators continues even when no early mediators are present. & \textsc{Source}$(l)$ & $f_{l}(\mathbf{0}) > 0$ \\
11 & The pathogen does not induce late mediators directly; it does so only by activating the early mediators. & \textsc{Absent}$(l,p)$ & $\partial f_{l}/\partial p \equiv 0$ \\
12 & Late mediators are cleared at a steady rate, so their level falls faster when more of them are present. & \textsc{Sign}$(l,l,-)$ & $\partial f_{l}/\partial l < 0$ \\
13 & All three quantities are amounts and remain non-negative. & \textsc{Pos}$(p)$ \newline \textsc{Pos}$(m)$ \newline \textsc{Pos}$(l)$ & $p=0 \Rightarrow f_{p} \ge 0$ \newline $m=0 \Rightarrow f_{m} \ge 0$ \newline $l=0 \Rightarrow f_{l} \ge 0$ \\
\bottomrule
\end{tabular}
\caption{\textsc{Kumar} \citep{Kumar2004Dynamics}: description and gold constraints
(13 sentences, 15 constraints; 2 sentences state no constraint).
States: $p$ (pathogen), $m$ (early pro-inflammatory mediators), $l$ (late pro-inflammatory mediators).}
\label{tab:gold-kumar}
\end{table*}
 
\begin{table*}[p]
\centering
\footnotesize
\renewcommand{\arraystretch}{1.15}
\begin{tabular}{@{}r p{0.52\textwidth} p{0.17\textwidth} p{0.21\textwidth}@{}}
\toprule
\# & Sentence of the description & Gold constraint & Predicate \\
\midrule
1 & The pro-inflammatory mediator attracts neutrophils into the tissue, so more mediator makes the number of neutrophils rise faster. & \textsc{Sign}$(n,c,+)$ & $\partial f_{n}/\partial c > 0$ \\
2 & Neutrophils are short-lived: they undergo apoptosis, which removes them from the pool of live neutrophils. & \textsc{Sign}$(n,n,-)$ & $\partial f_{n}/\partial n < 0$ \\
3 & Each neutrophil that undergoes apoptosis becomes an apoptotic neutrophil, so more live neutrophils make apoptotic neutrophils accumulate faster. & \textsc{Sign}$(a,n,+)$ & $\partial f_{a}/\partial n > 0$ \\
4 & Macrophages do not remove live neutrophils. & \textsc{Absent}$(n,m)$ & $\partial f_{n}/\partial m \equiv 0$ \\
5 & Apoptotic neutrophils that are not removed in time break down, which removes them from the apoptotic pool. & \textsc{Sign}$(a,a,-)$ & $\partial f_{a}/\partial a < 0$ \\
6 & When they break down they release their contents, and this produces more of the pro-inflammatory mediator, so inflammation can feed itself. & \textsc{Sign}$(c,a,+)$ & $\partial f_{c}/\partial a > 0$ \\
7 & This release reaches a ceiling: when apoptotic neutrophils are already abundant, more of them add little extra mediator. & \textsc{Sat}$(c,a)$ & $\lim_{a\to\infty} f_{c}$ finite \\
8 & Live neutrophils and macrophages do not themselves produce the mediator. & \textsc{Absent}$(c,n)$ \newline \textsc{Absent}$(c,m)$ & $\partial f_{c}/\partial n \equiv 0$ \newline $\partial f_{c}/\partial m \equiv 0$ \\
9 & The mediator is cleared at a steady rate, so its level falls when nothing produces it. & \textsc{Sign}$(c,c,-)$ & $\partial f_{c}/\partial c < 0$ \\
10 & The mediator also attracts macrophages, so more mediator makes the number of macrophages rise faster. & \textsc{Sign}$(m,c,+)$ & $\partial f_{m}/\partial c > 0$ \\
11 & Macrophages engulf apoptotic neutrophils on contact, so more macrophages remove apoptotic neutrophils faster, before they can break down. & \textsc{Sign}$(a,m,-)$ \newline \textsc{Inter}$(a;a,m)$ & $\partial f_{a}/\partial m < 0$ \newline a term of $f_{a}$ has $a$ and $m$ \\
12 & Macrophages leave the tissue at a steady rate, so their number falls when the mediator no longer attracts new ones. & \textsc{Sign}$(m,m,-)$ & $\partial f_{m}/\partial m < 0$ \\
13 & All four quantities are amounts and remain non-negative. & \textsc{Pos}$(n)$ \newline \textsc{Pos}$(c)$ \newline \textsc{Pos}$(a)$ \newline \textsc{Pos}$(m)$ & $n=0 \Rightarrow f_{n} \ge 0$ \newline $c=0 \Rightarrow f_{c} \ge 0$ \newline $a=0 \Rightarrow f_{a} \ge 0$ \newline $m=0 \Rightarrow f_{m} \ge 0$ \\
\bottomrule
\end{tabular}
\caption{\textsc{Dunster} \citep{Dunster2014Resolution}: description and gold constraints
(13 sentences, 18 constraints; 0 sentences state no constraint).
States: $n$ (neutrophils), $c$ (pro-inflammatory mediator), $a$ (apoptotic neutrophils), $m$ (macrophages).}
\label{tab:gold-dunster}
\end{table*}
 
\begin{table*}[p]
\centering
\footnotesize
\renewcommand{\arraystretch}{1.15}
\begin{tabular}{@{}r p{0.52\textwidth} p{0.17\textwidth} p{0.21\textwidth}@{}}
\toprule
\# & Sentence of the description & Gold constraint & Predicate \\
\midrule
1 & Pro-inflammatory cytokines are produced at a low background rate even when none are present. & \textsc{Source}$(p)$ & $f_{p}(\mathbf{0}) > 0$ \\
2 & They also stimulate their own production, so inflammation can amplify itself, but this self-stimulation levels off when they are abundant. & -- & -- \\
3 & Anti-inflammatory cytokines inhibit the production of pro-inflammatory cytokines, so more anti-inflammatory cytokines make pro-inflammatory cytokines rise more slowly. & \textsc{Sign}$(p,a,-)$ & $\partial f_{p}/\partial a < 0$ \\
4 & This inhibition acts on both the background production and the self-stimulated production. & \textsc{Inter}$(p;a,p)$ & a term of $f_{p}$ has $a$ and $p$ \\
5 & Pro-inflammatory cytokines are cleared at a steady rate. & -- & -- \\
6 & Pro-inflammatory cytokines stimulate the production of anti-inflammatory cytokines, so more pro-inflammatory cytokines make anti-inflammatory cytokines rise faster. & \textsc{Sign}$(a,p,+)$ & $\partial f_{a}/\partial p > 0$ \\
7 & This production reaches a ceiling: when pro-inflammatory cytokines are already abundant, more of them add little extra production. & \textsc{Sat}$(a,p)$ & $\lim_{p\to\infty} f_{a}$ finite \\
8 & Anti-inflammatory cytokines are cleared at a steady rate, so their level falls when nothing stimulates their production. & \textsc{Sign}$(a,a,-)$ & $\partial f_{a}/\partial a < 0$ \\
9 & Both quantities are amounts and remain non-negative. & \textsc{Pos}$(a)$ \newline \textsc{Pos}$(p)$ & $a=0 \Rightarrow f_{a} \ge 0$ \newline $p=0 \Rightarrow f_{p} \ge 0$ \\
\bottomrule
\end{tabular}
\caption{\textsc{Baker} \citep{Baker2013CytokineInflammation}: description and gold constraints
(9 sentences, 8 constraints; 2 sentences state no constraint).
States: $a$ (anti-inflammatory cytokines), $p$ (pro-inflammatory cytokines).}
\label{tab:gold-baker}
\end{table*}

\section{More details on the Agentic Workflow}

\subsection{Implementation of symbolic constraint verification}
\label{app:symbolic_verification}

Symbolic verification is implemented using \texttt{SymPy}
\cite{meurerSympy}. Let $f_i(\mathbf{x};\boldsymbol{\theta})$ denote
the right-hand side of the $i$th ODE, with states $x_j \geq 0$ and
parameters $\theta_k>0$. For a SIGN constraint on the effect of $x_j$ on
$f_i$, we compute and simplify the partial derivative
$\partial f_i/\partial x_j$ and determine whether its sign agrees with the
required activation or inhibition over the positive domain; if SymPy cannot
establish the sign symbolically, we use the 400-point numerical check
described in Section~\ref{sec:methods}. For ABSENT, we compute
$\partial f_i/\partial x_j$ and require the simplified expression to be
identically zero. For SAT, we evaluate
$\lim_{x_j\rightarrow\infty} f_i$ symbolically and require the resulting
limit to be finite. For INTER, we expand $f_i$ into additive terms and
inspect whether at least one term contains both specified state variables;
this is an exact syntactic test under our definition of interaction. For
SOURCE, all state variables are substituted with zero and the resulting
expression is simplified and required to be positive. Finally, for POS of
state $x_i$, we substitute $x_i=0$ while leaving the other states symbolic
and verify that the resulting boundary derivative satisfies
$f_i|_{x_i=0}\geq0$. Thus, the verifier operates directly on the symbolic
candidate equations rather than using trajectory data or an LLM judgment.

\subsection{Weak-form parameter estimation}
\label{app:weak_form}

After symbolic verification, the remaining candidate ODE structures
contain unknown numerical parameters that must be estimated from the
observed trajectories. We fit the parameters of each ODE independently.
For the $i$th state variable, let the candidate equation be

\begin{equation}
    \dot{x}_i(t)
    =
    f_i(\mathbf{x}(t);\boldsymbol{\theta}_i),
    \label{eq:strong_form}
\end{equation}

where $\mathbf{x}(t)=(x_1(t),\ldots,x_d(t))$ denotes the system state
and $\boldsymbol{\theta}_i$ contains only the unknown parameters of
equation $i$. Although other state variables may appear on the
right-hand side of $f_i$, the parameters of their governing equations
are not optimized jointly with $\boldsymbol{\theta}_i$.

\paragraph{Why use a weak formulation?}
A direct approach to fitting Eq.~\ref{eq:strong_form} would first
estimate $\dot{x}_i(t)$ from the measured trajectory and then minimize
the discrepancy between the estimated derivative and
$f_i(\mathbf{x}(t);\boldsymbol{\theta}_i)$. This is undesirable in our
setting because numerical differentiation amplifies measurement noise.
If the observed trajectory is

\begin{equation}
    y_i(t)=x_i(t)+\epsilon_i(t),
\end{equation}

then differentiating the observations produces

\begin{equation}
    \dot{y}_i(t)
    =
    \dot{x}_i(t)+\dot{\epsilon}_i(t),
\end{equation}

so even relatively small high-frequency perturbations in
$\epsilon_i(t)$ can lead to large errors in the estimated derivative.
This is particularly problematic for the noisy biological trajectories
considered in our experiments.

We therefore use a weak-form formulation following
Messenger and Bortz \cite{Messenger2021WeakSINDy}. Rather than differentiating
the observations, the ODE is integrated against a smooth test function.
Let $\phi_k(t)$ be a differentiable test function with compact support
on an interval $[a_k,b_k]$ and satisfying

\begin{equation}
    \phi_k(a_k)=\phi_k(b_k)=0.
\end{equation}

Multiplying Eq.~\ref{eq:strong_form} by $\phi_k(t)$ and integrating gives

\begin{equation}
    \int_{a_k}^{b_k}
    \phi_k(t)\dot{x}_i(t)\,dt
    =
    \int_{a_k}^{b_k}
    \phi_k(t)
    f_i(\mathbf{x}(t);\boldsymbol{\theta}_i)\,dt.
    \label{eq:weak_before_parts}
\end{equation}

Applying integration by parts to the left-hand side,

\begin{equation}
\begin{split}
    \int_{a_k}^{b_k}\phi_k(t)\dot{x}_i(t)\,dt
    &= \left[
        \phi_k(t)x_i(t)
    \right]_{a_k}^{b_k}
    \\
    & - \int_{a_k}^{b_k}
        \dot{\phi}_k(t)x_i(t)\,dt.
\end{split}
\end{equation}

Because $\phi_k(a_k)=\phi_k(b_k)=0$, the boundary term vanishes,
yielding

\begin{equation}
    -
    \int_{a_k}^{b_k}
        \dot{\phi}_k(t)x_i(t)\,dt
    =
    \int_{a_k}^{b_k}
        \phi_k(t)
        f_i(\mathbf{x}(t);\boldsymbol{\theta}_i)\,dt.
    \label{eq:weak_exact}
\end{equation}

Replacing the unknown states by the measured trajectories
$\mathbf{y}(t)$ gives the empirical weak-form relation

\begin{equation}
    -
    \int_{a_k}^{b_k}
        \dot{\phi}_k(t)y_i(t)\,dt
    \approx
    \int_{a_k}^{b_k}
        \phi_k(t)
        f_i(\mathbf{y}(t);\boldsymbol{\theta}_i)\,dt.
    \label{eq:weak_empirical}
\end{equation}

The important consequence is that differentiation is transferred from
the noisy trajectory $y_i(t)$ to the known smooth function $\phi_k(t)$.
The observations therefore enter the objective only through numerical
integration, which is substantially less sensitive to measurement noise
than pointwise numerical differentiation.

\paragraph{Weak-form residual.}
For each test function $\phi_k$, we define the residual

\begin{equation}
\begin{split}
r_{ik}(\boldsymbol{\theta}_i)
&=
-
\int_{a_k}^{b_k}
\dot{\phi}_k(t)y_i(t)\,dt
\\
&\quad -
\int_{a_k}^{b_k}
\phi_k(t)
f_i(\mathbf{y}(t);\boldsymbol{\theta}_i)\,dt.
\end{split}
\label{eq:weak_residual}
\end{equation}

The parameters of equation $i$ are then estimated independently by

\begin{equation}
    \hat{\boldsymbol{\theta}}_i
    =
    \arg\min_{\boldsymbol{\theta}_i>0}
    \sum_{k=1}^{K}
    r_{ik}(\boldsymbol{\theta}_i)^2.
    \label{eq:weak_objective}
\end{equation}

Using multiple test functions produces multiple integral constraints
from different portions of the trajectory and therefore provides an
overdetermined system from which the parameters can be estimated.

\paragraph{Parameters entering linearly.}
When the candidate equation is linear in its unknown parameters, it can
be written as

\begin{equation}
    f_i(\mathbf{x};\boldsymbol{\theta}_i)
    =
    \sum_{\ell=1}^{p_i}
    \theta_{i\ell}g_{i\ell}(\mathbf{x}),
    \label{eq:linear_parameterization}
\end{equation}

where the basis functions $g_{i\ell}$ are fixed by the symbolic
structure proposed in Stage~2. Substituting
Eq.~\ref{eq:linear_parameterization} into the weak form gives

\begin{equation}
    b_{ik}
    =
    \sum_{\ell=1}^{p_i}
    G_{ik\ell}\theta_{i\ell},
\end{equation}

where

\begin{equation}
    b_{ik}
    =
    -
    \int_{a_k}^{b_k}
    \dot{\phi}_k(t)y_i(t)\,dt
\end{equation}

and

\begin{equation}
    G_{ik\ell}
    =
    \int_{a_k}^{b_k}
    \phi_k(t)
    g_{i\ell}(\mathbf{y}(t))\,dt.
\end{equation}

Stacking the $K$ test functions yields

\begin{equation}
    \mathbf{b}_i
    =
    \mathbf{G}_i\boldsymbol{\theta}_i.
\end{equation}

Because all free kinetic parameters are constrained to be positive in
P2C, we estimate them using non-negative least squares,

\begin{equation}
    \hat{\boldsymbol{\theta}}_i
    =
    \arg\min_{\boldsymbol{\theta}_i\geq0}
    \left\|
        \mathbf{G}_i\boldsymbol{\theta}_i-\mathbf{b}_i
    \right\|_2^2.
    \label{eq:nnls}
\end{equation}

Importantly, an ODE may be nonlinear in its state variables while still
being linear in its parameters. For example,

\begin{equation}
    \dot{x}_i
    =
    \theta_{i1}x_i
    -
    \theta_{i2}x_ix_j
\end{equation}

is nonlinear in $(x_i,x_j)$ but linear in
$(\theta_{i1},\theta_{i2})$, and can therefore be fit using
Eq.~\ref{eq:nnls}.

\paragraph{Parameters entering nonlinearly.}
Some candidate structures contain parameters inside nonlinear
expressions. For example,

\begin{equation}
    \dot{x}_i
    =
    \theta_{i1}
    \frac{x_j}{\theta_{i2}+x_j}
    -
    \theta_{i3}x_i
\end{equation}

contains $\theta_{i2}$ inside a saturation term and cannot be reduced
to a linear least-squares problem. For such candidates, we minimize the
weak-form objective in Eq.~\ref{eq:weak_objective} directly using a
bounded numerical optimizer, constraining all parameters to remain
positive. Because the resulting objective may be non-convex, the
optimization is repeated from multiple random initializations and the
solution with the lowest weak-form residual is retained.

% \paragraph{Numerical implementation.}
% For each surviving candidate equation, the weak-form integrals are
% computed independently for a collection of $K$ test functions.
% The left-hand-side integrals depend only on the observed trajectory and
% can therefore be computed once for each state and test function, while
% the right-hand side is recomputed as the candidate parameters change.
% After fitting, the resulting parameterized ODE is passed to the model
% scoring stage, where it is simulated from the observed initial
% condition and evaluated using trajectory-level prediction error.
% Thus, weak-form residuals are used for robust parameter estimation,
% whereas the final model ranking is based on the behavior of the fitted
% dynamical system itself.

\paragraph{Numerical implementation.}
\emph{Test functions.} We use polynomial function
\begin{equation}
  \phi_k(t) =
  \begin{cases}
    Z_k^{-1}\,\bigl(1 - s_k(t)^2\bigr)^{4}, & |s_k(t)| < 1,\\
    0, & \text{otherwise},
  \end{cases}
  \label{eq:bump}
\end{equation}
where $s_k(t) = (t-c_k)/h_k$, and with derivative
\begin{equation}
  \dot\phi_k(t) = -\frac{8\,s_k(t)}{Z_k h_k}\bigl(1 - s_k(t)^2\bigr)^{3}
  \label{eq:bump-deriv}
\end{equation}
on the support. The power four makes $\phi_k$ and $\dot\phi_k$ vanish at both
ends of the support, so the boundary term in Eq.~(10) is zero. $Z_k$
normalises each bump to unit integral, so that all test functions carry
equal weight. The half-widths are $h_k \in \{0.04, 0.08, 0.16\}\,T$, where
$T$ is the length of the time window (supports of 8\%, 16\% and 32\% of
the window), and for each width the centres $c_k$ run from $h_k$ to
$T-h_k$ in steps of $h_k/2$. This gives $47+22+9=78$ test functions per
regime and state, or 234 weak-form equations per state over the three
training regimes.

\emph{Quadrature.} Both integrals in Eq.~(12) are computed with the
composite trapezoid rule on the fitting points of each regime (the 101
points with even index, spacing $T/100$). The left-hand side
$b_{ik}=-\int\dot\phi_k y_i\,dt$ is computed once per candidate; the
right-hand side is recomputed for every parameter value. Observations are
clipped at zero before integration.

\emph{Scaling.} The residuals of equation $i$ are divided by the root mean
square of $b_{ik}$ over all test functions and regimes, so that every state
contributes equally regardless of its magnitude.

\emph{Linear parameters.} Parameters that multiply a whole term are
solved, for each equation separately and for every value of the nonlinear
parameters, by non-negative least squares after normalising each column of $\mathbf G_i$ to unit norm. This
is a variable-projection scheme: the outer optimiser sees only the
nonlinear parameters.

\emph{Nonlinear parameters.} The remaining parameters are optimised jointly
over all equations in log space, $\theta=\exp(\eta)$, with SciPy's
trust-region reflective least-squares solver, bounds
$\theta\in[10^{-4},10^{4}]$ and at most 200 residual evaluations per start.
We use 24 starts drawn uniformly in $\log\theta\in[\log 10^{-2},\log 10^{2}]$
from a seeded generator, and keep the start with the lowest mean squared
scaled residual. A candidate without nonlinear parameters is solved in one
step.

\emph{Trajectory refinement.} The weak-form estimate initialises a
refinement of all parameters jointly, again in log space, with the same
solver, bounds of $\pm 6$ around the initial log values, at most 60
residual evaluations and a finite-difference step of $10^{-3}$. The
residual is $(\hat x_i(t)-y_i(t))/\mathrm{RMS}(y_i)$ over all fitting points,
states and training regimes, where $\hat x$ is simulated with LSODA
(relative tolerance $10^{-6}$, absolute tolerance $10^{-9}$); a failed
simulation contributes a residual of 10 at every point. A fit that takes
longer than 240\,s is stopped and the candidate is treated as failed.

\emph{Measured data (\S\ref{sec:app-exp1}).} For series with fewer than
30 points the half-widths are two and three sampling intervals instead,
which gives five test functions per eight-point series.

\subsection{Baseline implementation and comparison protocol}
\label{app:baseline-comparison}

The comparisons with LLM-SR, DoLQ, and EDL evaluate the complete
discovery systems rather than isolating a single algorithmic
component. The methods differ in their candidate-generation
strategies, search dynamics, and parameter-estimation procedures, as
these components are intrinsic to the respective methods. We
therefore do not interpret differences between P2C and these baselines
as being caused solely by the text-derived constraints.

To make the comparison as controlled as possible, all methods receive
the same state variables, biological description, training
trajectories, and noise realization for each system--noise--seed
configuration. The held-out regime and the corresponding noise-free
trajectory are used only for evaluation. We use the same LLM backbone
for all LLM-based methods unless otherwise stated.

P2C uses its weak-form parameter estimator described above, whereas the baselines retain their
method-specific fitting and search procedures. This choice evaluates
each method as a complete equation-discovery framework, but means that
the cross-method comparison reflects differences in both structural search and numerical estimation.

\subsection{Computational Resources}
\label{app:compute}

P2C trains no model and uses no GPU. All LLM calls, namely Stage~1
extraction ($K=3$ runs per system) with its entailment checks, the
Stage~2 proposals, and the LLM-SR, DoLQ and EDL baselines, use
GPT-6-Luna through the OpenAI API with its default sampling settings;
its parameter count is not publicly disclosed. The only parameters
estimated in this work are the coefficients of the candidate ODEs,
fitted for each candidate as described in Appendix~\ref{app:weak_form}.
Stage~1 runs once per system. A P2C run then makes $N=25$ proposal
calls with three candidates each, so it fits at most 75 candidates,
each fit limited to 240\,s. Symbolic verification, parameter fitting
and simulation run on CPU in Google Colab ([CPU: Intel(R) Xeon(R) CPU @ 2.20GHz, RAM: 13.61]). The benchmark uses ten seeds for each (system, noise level, method) cell, and the application uses three seeds per method; in total the experiments used about 500 CPU-hours and 2250 API calls.

\section{Application: data, descriptions and evaluation}
\label{app:application}
 
This appendix gives the material needed to repeat Experiment~1
(\S\ref{sec:app-exp1}): the data (\S\ref{app:app-data}), the prompt and
the description written by the LLM (\S\ref{app:prose}), the changes
to the pipeline for measured data, the evaluation (\S\ref{app:app-eval}) and all runs (\S\ref{app:app-runs}).
 
% ---------------------------------------------------------------------
\subsection{Data}
\label{app:app-data}
 
Table~\ref{tab:app-data} lists the series. The values are the group
medians of Fig.~3c of \citet{Zachs2019Noninvasive}, read from the box plots: the
increase of ankle thickness above day~0, in mm. On day~0 the mice receive
K/BxN serum; from then on each group receives ultrasound of the spleen
(1~MHz, 350~kPa) for a fixed number of minutes per day. The sham group has
the probe in place with the ultrasound switched off. The input $u$ is the
daily duration, and $u=0$ for the sham group. The initial condition of every
series is the measured value on day~0, and time is in days.
 
\begin{table}[t]
\centering
\small
\setlength{\tabcolsep}{3.5pt}
\begin{tabular}{@{}rr*{8}{c}@{}}
\toprule
$u$ & $n$ & d0 & d1 & d2 & d3 & d4 & d5 & d6 & d7 \\
\midrule
0  & 16 & 0 & .011 & .228 & .451 & .552 & .777 & .987 & 1.112 \\
2  & 16 & 0 & .005 & .104 & .277 & .379 & .603 & .703 & .727 \\
6  & 6  & 0 & 0    & 0    & .043 & .144 & .192 & .279 & .354 \\
12 & 6  & 0 & 0    & .031 & .104 & .167 & .279 & .367 & .478 \\
20 & 6  & 0 & 0    & 0    & .072 & .056 & .079 & .180 & .205 \\
\bottomrule
\end{tabular}
\caption{Median ankle swelling (mm) by day. $u$: minutes of splenic ultrasound per day ($0$ = sham); $n$: mice per group. $u=20$ group is held out for validation.}
\label{tab:app-data}
\end{table}
 
% ---------------------------------------------------------------------
\subsection{Description written by the LLM}
\label{app:prose}
 
\paragraph{Prompt.}
The description is written once by the LLM and is then given unchanged to
every method and every seed. The prompt below gives the experiment, the
model variables and seven notes on the cholinergic anti-inflammatory
pathway. The notes are short statements from the literature
(Table~\ref{tab:app-notes}); none of them says how the swelling changes
with the duration of stimulation, so this relation must be reasoned from
the pathway. The prompt first asks the LLM to follow the pathway link by
link and to answer five questions per link, which correspond to the
constraint types \textsc{Sign}, \textsc{Sat}, \textsc{Source},
\textsc{Pos} and \textsc{Absent}; it then asks for 8--12 sentences
without equations, numbers, names of functions or results of the
experiment. The prompt is shown with its slots filled in.
 
\begin{table}[t]
\centering
\small
\begin{tabular}{@{}lp{0.72\columnwidth}@{}}
\toprule
Note & Source \\
\midrule
1    & \citet{Zachs2019Noninvasive} \\
2    & \citet{Tracey2002InflammatoryReflex}; \citet{RosasBallina2011Acetylcholine}; \citet{Wang2003Nicotinic} \\
3, 4 & \citet{Cotero2019Noninvasive} \\
5    & \citet{Levine2014Neurostimulation}; \citet{Koopman2016Vagus} \\
6, 7 & \citet{Zachs2019Noninvasive}; \citet{Cotero2019Noninvasive} \\
\bottomrule
\end{tabular}
\caption{Sources of the notes in the description prompt.}
\label{tab:app-notes}
\end{table}
 
{\scriptsize
\begin{verbatim}
You are a biomedical scientist. You write a short
mechanistic description of an experiment.
A second program reads your description sentence by
sentence and turns it into formal constraints on a
differential-equation model. So each sentence must say
one clear thing about the variables of the model.
 
EXPERIMENT
Mice receive an injection of serum from arthritic donor
mice on day 0 (the K/BxN serum-transfer model of
inflammatory arthritis). In the following days their
ankles swell. Each group of mice receives noninvasive
ultrasound stimulation of the spleen every day, for a
fixed number of minutes per day that differs between the
groups. One group receives no stimulation. All other
ultrasound settings are the same in all groups.
 
VARIABLES OF THE MODEL
A: ankle swelling (the increase of the ankle thickness
above its value before the serum transfer, in
millimetres)
u (known input, a positive number that differs between
experiments): daily duration of the ultrasound
stimulation of the spleen, in minutes
Time is in days. The input u is zero in the group
without stimulation.
 
NOTES FROM THE LITERATURE
1. In the serum-transfer model, the transferred
   antibodies cause an inflammation of the joints. The
   ankles and paws swell during the week after the
   transfer. The measurements cover this week.
2. The cholinergic anti-inflammatory pathway: nerve
   signals reach the spleen through the splenic nerve,
   which releases norepinephrine. A group of T cells in
   the spleen then releases acetylcholine. Acetylcholine
   acts on the alpha-7 nicotinic receptor of macrophages
   and suppresses their release of inflammatory
   cytokines such as TNF.
3. Ultrasound that is focused on the spleen activates
   this pathway without surgery. It raises
   norepinephrine and acetylcholine in the spleen and
   lowers the TNF response. The effect is lost in
   animals without acetylcholine-producing T cells and
   in animals without the alpha-7 receptor.
4. After one session of ultrasound, the cytokine
   response stays lowered for many hours.
5. Stimulation of this pathway lowers cytokine
   production and disease severity in arthritis, in
   animals and in patients.
6. Ultrasound of other body sites, for example the leg
   or the side opposite to the spleen, does not have
   this effect.
7. With the settings of this experiment, the ultrasound
   causes no relevant heating of the tissue.
 
STEP 1. FOLLOW THE PATHWAY
Go from the known input to each state variable, one link
at a time: what the stimulus acts on, which
signals and cells carry the effect, what they do to the
process that drives the state variable, and what
the state variable does when there is no stimulus. For
each link, answer these questions:
(a) Direction: does more of the cause make the state
    variable rise faster or more slowly?
(b) Limit: can the effect grow without bound, or does
    the mechanism give it a ceiling? Say why.
(c) Baseline: does the state variable change when all
    variables of the model are zero?
(d) Range: can the state variable become negative?
(e) Absence: is there an effect that the notes deny?
 
STEP 2. WRITE THE DESCRIPTION
Write 8 to 12 sentences in plain English.
1. One sentence states one mechanism or one property.
   Use the plain names of the variables, not the
   symbols.
2. Say how a cause changes the RATE at which a state
   variable rises or falls (for example "makes it build
   up more slowly"). Do not describe only the final
   level.
3. State direction, ceiling, baseline production, non-
   negativity and denied effects in separate sentences,
   if the pathway supports them. If the pathway does not
   decide a point, leave it out. Do not guess.
4. Do not write equations, symbols, numbers, units, or
   names of functions (for example "linear",
   "exponential", "proportional", "Hill").
5. Do not report results of this experiment (no
   comparison of groups, no statistics). Describe the
   mechanism that is expected to hold.
6. Do not name papers or authors.
 
OUTPUT
JSON only: {"pathway": ["<link 1>", "<link 2>", "..."],
"description": "<the sentences as one paragraph>"}
\end{verbatim}
}
 
\paragraph{Output.}
With GPT-6-Luna the LLM returned the following pathway and description.
Both are reproduced without changes. The description passed the automatic
check for numbers, equations and names of functions.
 
\begin{quote}
\small
\textit{Pathway.}
(1)~Spleen-focused ultrasound activates splenic nerve signaling and
increases norepinephrine in the spleen.
(2)~Norepinephrine stimulates acetylcholine-producing T cells in the spleen
to release acetylcholine.
(3)~Acetylcholine activates alpha-seven nicotinic receptors on macrophages.
(4)~Receptor activation suppresses macrophage release of inflammatory
cytokines, including TNF.
(5)~Reduced cytokine production slows joint inflammation, which slows the
rise in ankle swelling.
(6)~Transferred antibodies drive joint inflammation and ankle swelling even
without ultrasound stimulation.
 
\textit{Description.}
Spleen-focused ultrasound activates splenic nerve signaling and increases
norepinephrine in the spleen. Norepinephrine stimulates
acetylcholine-producing T cells in the spleen to release acetylcholine.
Acetylcholine activates alpha-seven nicotinic receptors on macrophages.
Receptor activation suppresses macrophage release of TNF and other
inflammatory cytokines. Reduced cytokine production makes joint
inflammation build more slowly. Slower joint inflammation makes ankle
swelling rise more slowly. Transferred antibodies continue to drive joint
inflammation and ankle swelling without ultrasound stimulation. The
ultrasound-induced cytokine suppression depends on acetylcholine-producing
T cells and macrophage alpha-seven nicotinic receptors. The cytokine
response remains lowered for many hours after a session. Ultrasound at
other body sites does not produce this anti-inflammatory effect. These
ultrasound settings do not cause relevant tissue heating.
\end{quote}
 
% ---------------------------------------------------------------------
% \subsection{Reference description and reference constraints}
% \label{app:app-reference}
 
% The reference description (Table~\ref{tab:app-reference}) was written by
% hand under the same rules as the descriptions of the benchmark. Four of its sentences state a constraint. These four constraints are the reference set that is used to
% count violations in Table~\ref{tab:app-exp1} and to score Stage~1. They are
% statements about the expected mechanism, not ground truth: the
% \textsc{Sat} constraint in particular is an inference from the mechanism
% (the pathway can at most switch off the cytokine release) and is not
% tested by the data. In Experiment~1 the methods receive the description
% written by the LLM, not the reference description.
 
\begin{table*}[t]
\centering
\small
\begin{tabular}{@{}p{0.47\textwidth}llp{0.16\textwidth}@{}}
\toprule
Sentence & Constraint & Predicate on $f$ & Source \\
\midrule
The mice receive serum from arthritic donor mice, and the antibodies in
this serum start an inflammation of the joints.
  & -- & -- & \citet{Zachs2019Noninvasive} \\
This inflammation makes the ankles swell, so ankle swelling builds up
after the serum transfer even when there is no swelling yet.
  & \textsc{Source}$(A)$ & $f(0,u) > 0$ & \citet{Zachs2019Noninvasive} \\
Ankle swelling is measured as an increase above the healthy ankle, and it
does not fall below zero.
  & \textsc{Pos}$(A)$ & $A=0 \Rightarrow f \ge 0$ & measurement \\
Ultrasound stimulation of the spleen activates the cholinergic
anti-inflammatory pathway: the splenic nerve releases norepinephrine, and
T cells in the spleen then release acetylcholine.
  & -- & -- & \citet{Cotero2019Noninvasive, RosasBallina2011Acetylcholine} \\
Acetylcholine acts on macrophages and suppresses their release of
inflammatory cytokines, which are the signals that drive the joint
inflammation.
  & -- & -- & \citet{Tracey2002InflammatoryReflex} \\
A longer daily stimulation therefore makes ankle swelling build up more
slowly.
  & \textsc{Sign}$(A,u,-)$ & $\partial f/\partial u < 0$ & \citet{Zachs2019Noninvasive} \\
The pathway can only suppress the cytokine release and cannot do more than
switch it off, so the benefit of stimulation has a ceiling: however long
the daily stimulation, it cannot slow the build-up of swelling beyond that
ceiling.
  & \textsc{Sat}$(A,u)$ & $\lim_{u\to\infty} f$ finite & inference \\
Ultrasound applied to other parts of the body does not have this effect,
which shows that it acts through the spleen and not on the joint itself.
  & -- & -- & \citet{Zachs2019Noninvasive, Cotero2019Noninvasive} \\
\bottomrule
\end{tabular}
\caption{Reference description for Experiment~1, sentence by sentence,
with the constraint each sentence states. $\dot A = f(A,u)$. All predicates
are decided for $A>0$, $u>0$ and positive parameters
(Table~\ref{tab:schema}).}
\label{tab:app-reference}
\end{table*}

\subsection{Evaluation}
\label{app:app-eval}
 
\paragraph{Held-out error.}
For the held-out group, $\mathrm{NMSE} =
\frac{1}{T}\sum_t (\hat A_t - A_t)^2 / \operatorname{Var}(A)$, where $\hat A$
is the simulation of the returned equation from the measured value on day~0
and $T=8$. For the $u=20$ series, a model that predicts no swelling has an
NMSE of $1 + \bar A^2/\operatorname{Var}(A) = 1.97$; this value appears in
several rows of Table~\ref{tab:app-seeds}. 

\paragraph{Violated reference constraints.}
The generated equation is first simplified: a term whose size on the
training data is below $10^{-3}$ of the largest term of its equation is
removed. The checker of Stage~3 then tests the four reference constraints
of Table~\ref{tab:app-reference} on the simplified form, for all positive
parameter values. A run that returns no equation is counted as violating
all four.
 
\paragraph{Behaviour beyond the tested durations.}
We evaluate $f(0,u)$, the rate of change at zero swelling, at
$u \in \{40, 60, 120, 240\}$ minutes per day. If it is negative at any of
these durations, the model predicts that swelling falls below zero, which
the measurement does not allow; we report this as a \emph{negative rate}.
In Experiment~1 this holds for LLM-SR (seed~1), DoLQ (seed~1) and EDL
(seed~2), and for no P2C run.
 
% ---------------------------------------------------------------------
\subsection{All runs}
\label{app:app-runs}
 
Table~\ref{tab:app-seeds} lists the equation returned in every run.

\begin{table*}
\centering
\small
\begin{tabular}{@{}llccp{0.6\textwidth}@{}}
\toprule
Method & Seed & NMSE & Violations & Returned equation $\dot A$ \\
\midrule

P2C & 0 & 0.218 & 0 &
$0.153/(1+0.31u)$ \\

P2C & 1 & 0.218 & 1 &
$0.153(1-2.48\times10^{-7}A)/(1+0.31u)$ \\

P2C & 2 & 0.218 & 0 &
$0.153/(1+0.31u)$ \\

LLM-SR & 0 & \multicolumn{3}{l}{No equation returned} \\

LLM-SR & 1 & 1.97 & 3 &
$0.0674 + 0.0842A + 0.00991Au - 0.00389u$ \\

LLM-SR & 2 & \multicolumn{3}{l}{No equation returned} \\

DoLQ & 0 & 5.76 & 2 &
$-1.36A^2 + 0.212Ae^{-u^2} - 4.3A/(A^2+1)
 + 5.45 - 5.44e^{-A}$ \\

DoLQ & 1 & 1.97 & 3 &
$-0.316A^2 + 0.381A - 0.00285u + 0.0313$ \\

DoLQ & 2 & 6.96 & 0 &
$-0.21Au/(u^2+1) - 0.0286Au/(u+1)
 + 0.309A/(A+1) + 0.0361$ \\

EDL & 0 & 5.76 & 2 &
$0.00279u + 0.155e^{-0.133u}\tanh(83.9A)$ \\

EDL & 1 & 1.97 & 2 &
$0.179A^{0.28} - 63.9A^{18.7}\log(u+1)$ \\

EDL & 2 & 1.97 & 2 &
$-0.249A^{0.0441}/(e^{-0.589u}-2.34)-0.0164$ \\

\bottomrule
\end{tabular}

\caption{Returned equations and held-out performance for all seeds in
Experiment~1 ($u=20$). Parameters are rounded to three significant
figures. Lower NMSE is better.}
\label{tab:app-seeds}
\end{table*}

\end{document}